\documentclass[10pt,twocolumn,letterpaper]{article}

\usepackage[pagenumbers]{cvpr} 

\definecolor{cvprblue}{rgb}{0.21,0.49,0.74}
\usepackage[pagebackref,breaklinks,colorlinks,allcolors=cvprblue]{hyperref}
\newcommand{\arrowdown}{\ensuremath{\downarrow}}  
\newcommand{\arrowup}{\ensuremath{\uparrow}}      
\usepackage{hyperref}
\usepackage{graphicx}
\usepackage{multirow}
\usepackage[linesnumbered,ruled,vlined]{algorithm2e}
\SetKwInput{KwInput}{Input}
\SetKwInput{KwOutput}{Output}

\def\paperID{*****} 
\def\confName{CVPR}
\def\confYear{2026}

\title{dots.ocr: Multilingual Document Layout Parsing \\ in a Single Vision-Language Model}

\author{
Yumeng~Li, Guang~Yang, Hao~Liu, Bowen~Wang, Colin~Zhang\\
hi lab, Xiaohongshu Inc\\
\href{https://github.com/rednote-hilab/dots.ocr/tree/master}{%
  \raisebox{-0.15em}{\includegraphics[height=1em]{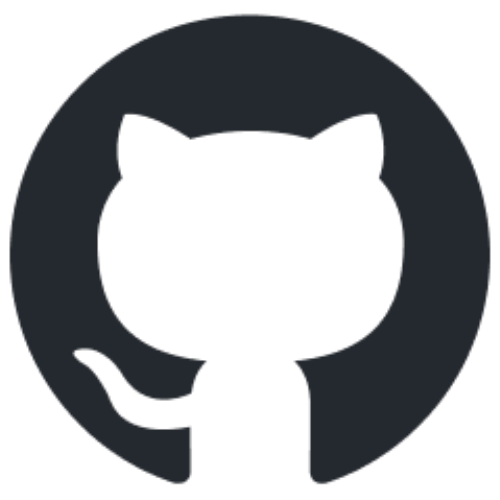}}~%
  \texttt{https://github.com/rednote-hilab/dots.ocr}
}
}

\begin{document}
\maketitle
\begin{abstract}
Document Layout Parsing serves as a critical gateway for Artificial Intelligence (AI) to access and interpret the world's vast stores of structured knowledge. This process, which encompasses layout detection, text recognition, and relational understanding, is particularly crucial for empowering next-generation Vision-Language Models. Current methods, however, rely on fragmented, multi-stage pipelines that suffer from error propagation and fail to leverage the synergies of joint training.
In this paper, we introduce dots.ocr, a single  Vision-Language Model that, for the first time, demonstrates the advantages of jointly learning three core tasks within a unified, end-to-end framework. This is made possible by a highly scalable data engine that synthesizes a vast multilingual corpus, empowering the model to deliver robust performance across a wide array of tasks, encompassing diverse languages, layouts, and domains.
The efficacy of our unified paradigm is validated by state-of-the-art performance on the comprehensive OmniDocBench. Furthermore, to catalyze research in global document intelligence, we introduce XDocParse, a challenging new benchmark spanning 126 languages. On this benchmark, dots.ocr achieves state-of-the-art performance, delivering an approximately 10\% relative improvement and demonstrating strong multilingual capability.

\end{abstract}    
\section{Introduction}
\label{sec:intro}

Document Layout Parsing \cite{smith2007overview,marinai2007machine,paliwal2019tablenet,xu2020layoutlm} is a fundamental capability for Artificial Intelligence to comprehend the vast repository of structured human knowledge. It not only powers downstream applications like intelligent process automation but also acts as a crucial data engine for training next-generation Vision-Language Model (VLM) \cite{singh2019towards,olivetti2020data}. Despite its foundational importance, parsing documents at scale, with high fidelity, and across the world's diverse spectrum of languages, layouts, and domains remains a formidable challenge.

\begin{figure*}[t]
  \centering
   \includegraphics[width=0.9\linewidth]{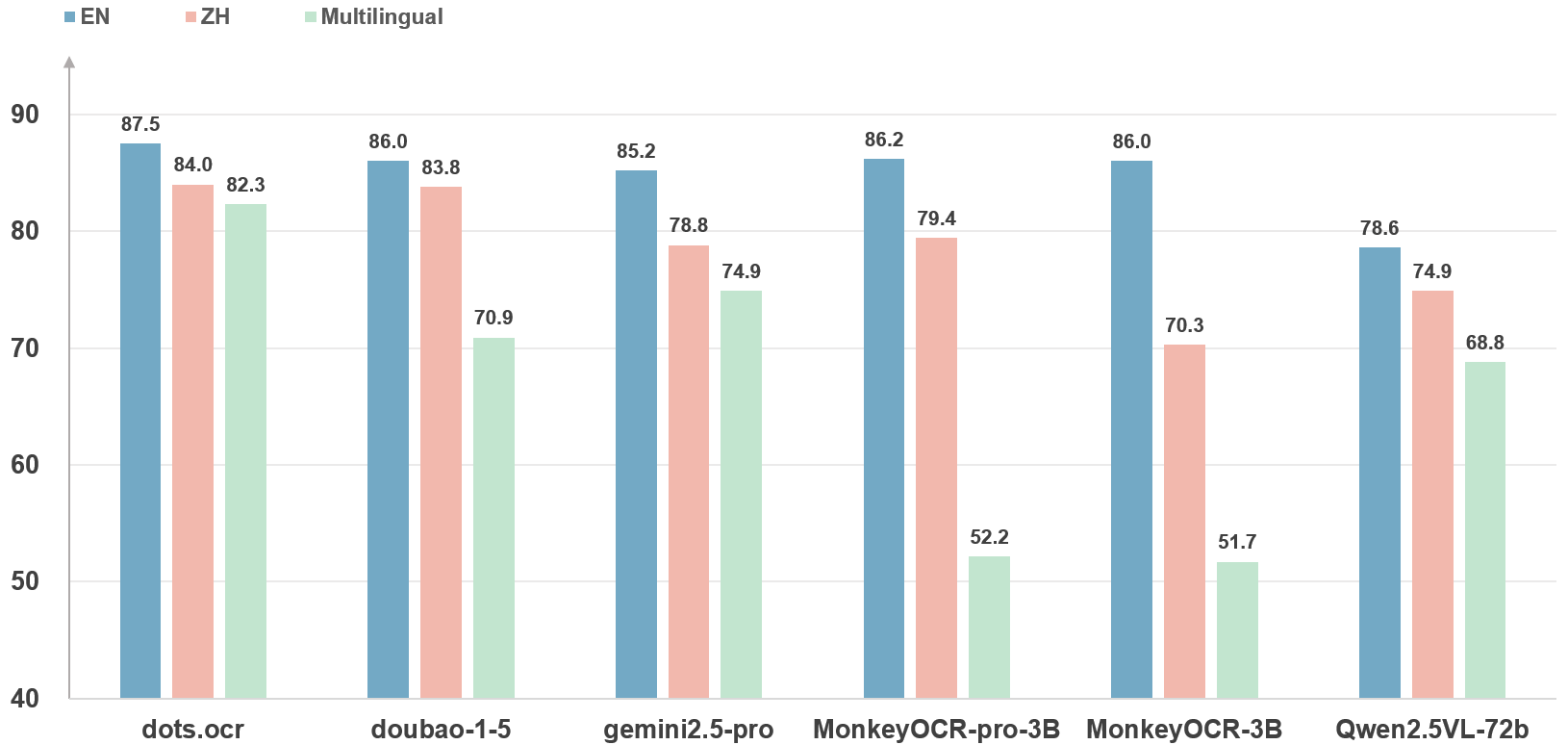}

   \caption{
    \textbf{Performance comparison of dots.ocr with other state-of-the-art methods.} 
    We evaluate performance on the comprehensive OmniDocBench\cite{ouyang2025omnidocbench} benchmark for English (EN) and Chinese (ZH), and on our newly proposed XDocParse benchmark for large-scale multilingual evaluation (Multilingual). 
    \textbf{dots.ocr} consistently outperforms all competing methods across all three settings, demonstrating its superior capability.
}
   \label{fig:score}
\end{figure*}

The first challenge is architectural. Comprehensive document parsing requires mastering a trinity of interconnected sub-tasks: (1) Layout Detection \cite{zhao2024detrs,huang2022layoutlmv3}, to identify the spatial boundaries of elements like paragraphs and figures; (2) Content Recognition \cite{du2025svtrv2,qu2023towards}, to extract the text and symbols within them; and (3) Relational Understanding \cite{sidorov2020textcaps,kuang2021mmocr,wang2021layoutreader}, to infer logical connections such as reading order. However, the dominant paradigm \cite{wang2024mineru,cui2025paddleocr} addresses these through fragmented, multi-stage pipelines, treating them as isolated steps. This separation, while modular, introduces two key limitations: it is prone to cascading errors, where mistakes in an early stage impairs subsequent ones, and fails to capitalize on the powerful synergies between tasks~\cite{zhang2023reading}.

A second major challenge concerns data and scale \cite{zhong2019publaynet,mathew2021docvqa}. Progress in document parsing has been overwhelmingly Anglo-centric, leaving a vast majority of the world's languages critically underserved \cite{xu2022xfund}. The root cause is the prohibitive cost of curating diverse, large-scale annotated datasets. This data bottleneck, in turn, creates a difficult trade-off between cost and performance: while pipeline systems are more data-efficient, they are complex to maintain and often hit a performance ceiling. Conversely, emerging end-to-end models \cite{poznanski2025olmocr, wang2024qwen2} promise higher performance but demand prohibitive computational resources, hindering their scalability.

To overcome these barriers, we introduce dots.ocr, a single, unified framework designed to achieve high performance, multilingual capability, and scalability. On the one hand, our unified VLM architecture is the first to demonstrate both the viability and superiority of jointly learning layout detection, content recognition, and relational understanding within a single end-to-end pass. On the other hand, to tackle the aforementioned data bottleneck, we built a highly scalable data engine. By synthesizing a vast corpus of high-quality multilingual training data, it alleviates the need for expensive manual annotation and enables robust generalization capabilities.

The effectiveness of our unified paradigm is rigorously demonstrated on the comprehensive OmniDocBench~\cite{ouyang2025omnidocbench}, where we achieve new state-of-the-art scores of 87.5 (EN) and 84.0 (CH). To further advance research in global document intelligence, we introduce XDocParse—a challenging new benchmark covering 126 languages. On this testbed, dots.ocr sets a robust new baseline, surpassing the previous best competitor by an impressive margin of +7.4 points, and showcasing its unparalleled multilingual capabilities.

In summary, our main contributions are as follows:
\begin{itemize}
\item \textbf{A Unified Architecture with State-of-the-Art Performance.} We introduce dots.ocr, a single VLM architecture that is the first to demonstrate both the viability and superiority of jointly learning layout detection, content recognition, and relational understanding. This holistic design, validated by extensive ablation studies showing its superiority over pipelined approaches, enables the model to deliver exceptional performance on OmniDocBench~\cite{ouyang2025omnidocbench}.


\item \textbf{A Scalable Multilingual Data Engine.} We designed a highly scalable data engine that powers our unified model. It systematically overcomes the data bottleneck by synthesizing a vast, diverse corpus, directly empowering the model with robust generalization capabilities across various languages, layouts, and document types. Futhermore, The critical role of this data engine is substantiated by extensive ablation studies.

\item \textbf{A New Large-Scale Multilingual Benchmark.} To spur research in global document intelligence, we introduce and will release XDocParse, a challenging benchmark spanning 126 languages. We establish a strong new baseline with dots.ocr, which significantly outperforms all existing methods on this demanding testbed.




\end{itemize}

\section{Related work}

\label{sec:2_related_work}
\subsection{General VLMs for Document Parsing}
The rapid advancement of powerful general Vision-Language Models (VLMs) like GPT-4o~\cite{hurst2024gpt}, Gemini 2.5 Pro~\cite{comanici2025gemini}, and Qwen2.5-VL~\cite{bai2025qwen2} has opened new avenues for high-level document understanding, such as summarization and question answering. However, their suitability for fine-grained, large-scale document parsing is limited by two fundamental challenges. First, as recent studies show~\cite{wang2024docllm}, these generalist models lack the architectural priors for structured analysis, often struggling with precise layout localization and dense text recognition. Second, their prohibitive computational cost and high latency render them impractical for processing the vast volumes of documents required in data pipelines. This fundamental mismatch, both in architecture and efficiency, necessitates the development of specialized VLMs explicitly designed for document parsing.
\subsection{Specialized VLMs for Document Parsing}
In response, a distinct line of research on specialized Document VLMs \cite{wan2024omniparser,peng2022spts} has emerged. The journey began with OCR-free pioneers like Donut~\cite{kim2021donut} and Nougat~\cite{blecher2023nougat}, which focused primarily on text recognition. More recently, the ambition has grown to unify the full spectrum of document parsing tasks, but current approaches remain fragmented in different ways. For instance, MonkeyOCR~\cite{li2025monkeyocr} relies on an explicit multi-stage pipeline, separately processing structure and content. Other models achieve a semblance of unification by omitting key tasks entirely. OlmOCR~\cite{poznanski2025olmocr}, for example, only parses content from given regions without layout detection. And Dolphin~\cite{feng2025dolphin}, which using a single model, internally employs a two-stage ``analyze-then-parse'' process. This staged design leads to suboptimal performance, particularly in complex tasks like reading order inference, revealing that a truly synergistic, unified paradigm has not yet been achieved.
In contrast, dots.ocr achieves synergistic joint learning by modeling the interdependencies between layout, content, and relations within a single, end-to-end generative process, which is a capability previously unattainable with staged approaches.

\subsection{Datasets for Multilingual Document parsing}

Beyond architectural design, a more fundamental challenge limiting the generalization of these specialized models is the severe scarcity of multilingual data for both training and evaluation~\cite{xia2024docgenome}. The vast majority of influential training datasets, such as M6Doc~\cite{cheng2023m6doc}, D4LA~\cite{da2023vision}, and DocLayNet~\cite{pfitzmann2022doclaynet}, are English-centric. Other resources, like CDLA~\cite{li2021cdla}, typically cover only one other major language. This lack of diverse training data hinders the development of truly global models~\cite{akter2023depth}. The problem is mirrored in the evaluation landscape: leading benchmarks like OmniDocBench~\cite{ouyang2025omnidocbench} also primarily focus on high-resource languages, making it impossible to assess true multilingual robustness. This dual scarcity in training and evaluation underscores the urgent need to actively generate diverse multilingual data and establish a comprehensive benchmark to fairly assess true cross-lingual generalization. 

\section{Method}
\label{sec:3_method}
Our approach, dots.ocr, is designed to perform end-to-end document parsing by unifying three core tasks within a single Vision-Language Model: layout detection, text recognition, and relational understanding, which we operationalize through the task of reading order generation. Realizing this unification requires fundamental innovations on two fronts, which we deliver through: (1) \textbf{a specialized training paradigm}, encompassing specific architectural choices and training strategies tailored for unified OCR tasks, and (2) \textbf{a holistic data engine} that generates the massive, multi-dimensionally diverse corpus necessary for robust training.




We now detail our approach. We begin by defining the unified task formulation that serves as the foundation for our method (Sec. \ref{sec:unified_formulation}). We then present the specifics of our model architecture and training process, detailing how we adapt a VLM for this unified task (Sec. \ref{sec:model_architecture}). Finally, we describe our holistic data engine, the mechanism responsible for generating the diverse, large-scale corpus that makes our unified training paradigm feasible (Sec. \ref{sec:data_engine}).

\subsection{Unified Task Formulation}
\label{sec:unified_formulation}

We formulate multilingual document parsing as a single, end-to-end autoregressive generation task. This unified formulation compels the model to learn the intricate interplay between an element's visual layout, textual content, and semantic role. 

Given a document image $\mathbf{I}$, the model outputs a structured sequence $\mathbf{S}$, where each entry corresponds to a semantic content block. Formally, the output sequence is:

\begin{equation}
    \mathbf{S} = \left [ (\mathcal{B}_1, c_1, t_1),\; \dots,\; (\mathcal{B}_K, c_K, t_K) \right ]
\end{equation}
where $K$ denotes the number of detected blocks. Each tuple $(\mathcal{B}_k, c_k, t_k)$ represents the $k$-th block, with $\mathcal{B}_k = (x_{k,1}, y_{k,1}, x_{k,2}, y_{k,2})$ as the bounding box coordinates, $c_k \in \mathcal{C}$ as the block category (e.g., title, header, table, figure), and $t_k$ as the recognized textual content. For unstructured blocks, $t_k$ is plain text; for structured regions such as tables, $t_k$ is encoded in Latex format to capture detailed layout and hierarchy.

Crucially, the sequence $\mathbf{S}$ is generated in an order that \textbf{aligns with human reading progression}. This task formulation compels the model to not only parse content but also to comprehend the document's logical flow, thus unifying layout detection, text recognition, and high-level relational understanding in a single pass.

\subsection{Model Architecture for Unified Parsing}
\label{sec:model_architecture}

\begin{figure*}[t]
  \centering
  \includegraphics[width=1.0\linewidth]{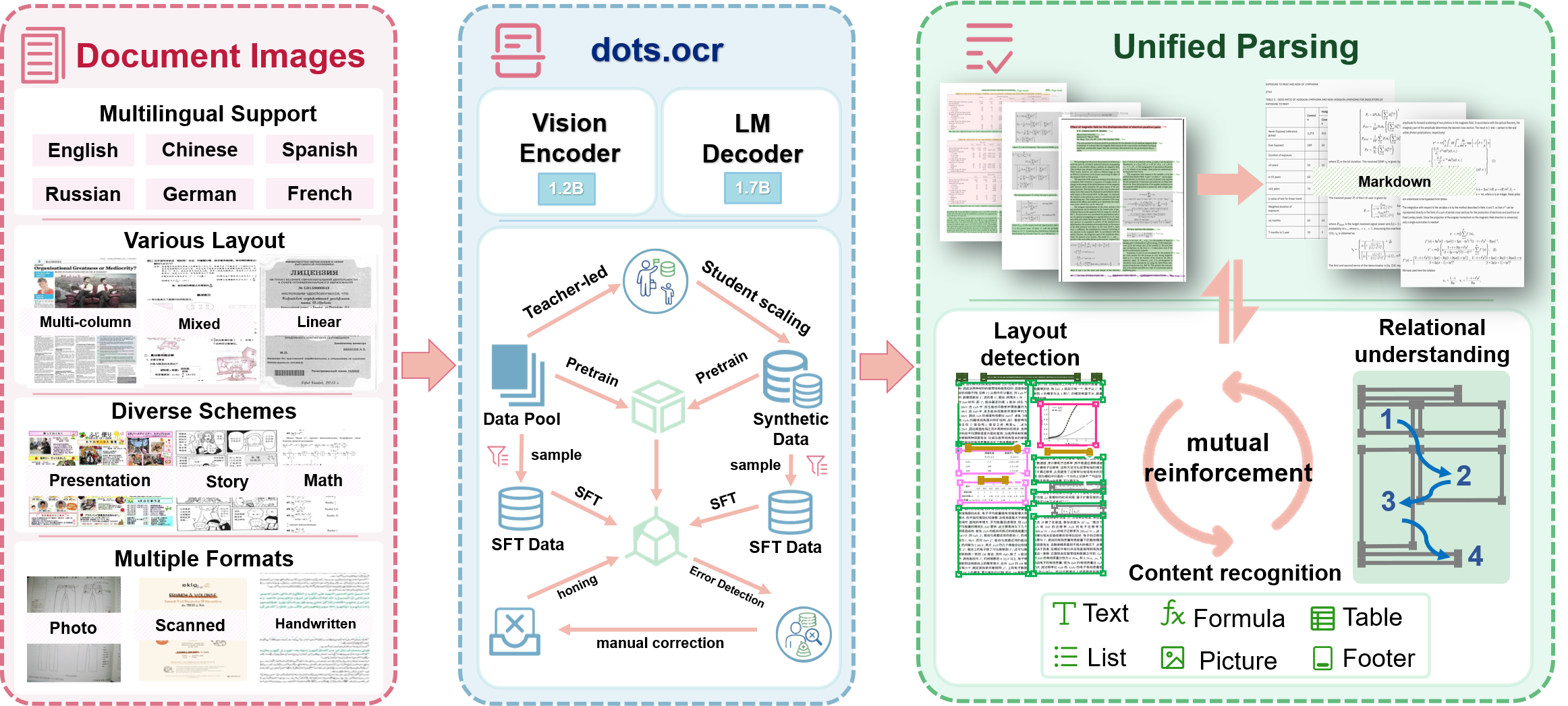}
    \caption{\textbf{Overview of the dots.ocr.} 
   \textbf{(Left)} Our system processes a diverse spectrum of document images, varying in language, layout, and format. 
   \textbf{(Center)} At its core is a Vision-Language Model (VLM) whose training is powered by our holistic Data Engine. 
   \textbf{(Right)} The model jointly performs three mutually reinforcing tasks—layout detection, content recognition, and relational understanding.}
   \label{fig:framework}
\end{figure*}
To accomplish our unified task, we adopt a ViT-LLM architecture. This design is inspired by recent advances in large vision-language models such as Qwen2-VL \cite{bai2025qwen2}, but incorporates critical modifications to both the vision encoder and the language decoder, tailored for unified document parsing.


\noindent \textbf{Vision Encoder:} We employ a 1.2B parameter Vision Encoder (VE) trained entirely from scratch, a deliberate departure from fine-tuning pre-trained, image-centric encoders. This from-scratch approach allows us to specialize its feature representation for document intelligence from the ground up. Architecturally, we design the VE to handle native high-resolution inputs of up to 11 million pixels \cite{dehghani2023patch}, enabling it to process dense documents. The training objective is multi-faceted, compelling the encoder to jointly master both fine-grained visual details for text recognition and high-level layout structures for relational understanding. This specialized, high-resolution VE provides a powerful foundation for the unified parsing task.


\noindent \textbf{Language Model Decoder:}
For the decoder, we select the Qwen2.5-1.5B base model as our foundation, a choice guided by the need to balance computational efficiency with the expressive capacity required for complex relational understanding. With modifications including tied word embeddings, the final decoder comprises 1.7B parameters. Crucially, initializing from the base model is a choice dictated by our large-scale pre-training paradigm. A base model provides the neutral, adaptable foundation essential for our training, which aims to teach the model the non-natural syntax and complex structures of diverse documents from the ground up, rather than adapting a model already specialized for conversational tasks.

\subsection{Data Engine for Unified Multilingual Parsing}
\label{sec:data_engine}

Training a unified model for multilingual document parsing imposes extreme demands on the training data. It requires a corpus that is not only massive in scale but also exceptionally diverse across several critical axes such as linguistic breadth, layout diversity, topical variety and so on. Since no existing dataset meets these stringent requirements, we engineered a novel, three-stage Data Engine to systematically construct our training corpus.

\textbf{Stage 1: Bootstrapping a Multilingual Synthesis Engine.}
The foundational challenge is the near-total absence of structured, labeled document data outside of high-resource languages. To overcome this, our first stage bootstraps a scalable synthesis engine via teacher-student distillation.


\textbf{(1) Teacher-Led Structured Generation.} We employ a powerful proprietary VLM (Qwen2.5-VL-72B\cite{bai2025qwen2}) as a ``teacher'' and task it with a structured re-rendering task: given a labeled English document and its structural representation, the teacher is prompted to generate a semantically equivalent document in a target language while preserving the original layout logic. This output is then rendered into a new image, creating a high-quality, parallel seed document.

\textbf{(2) Distilling to a Scalable Student.} This process, while high-quality, is too slow and expensive for mass production. Therefore, we use the generated multilingual seed corpus to fine-tune a smaller, more efficient Qwen2.5-VL-7B\cite{bai2025qwen2} model. This ``student'' model distills the core, structure-aware, cross-lingual generation capability of its teacher. The key outcome of this stage is not the data itself, but the creation of a specialized and highly scalable auto-labeling engine for the next stage.

\textbf{Stage 2: Strategically Curated Corpus for Large-Scale Pre-training.}
Leveraging our synthesis engine from Stage 1, we generate a massive pre-training corpus, prioritizing strategic curation over raw scale. We perform deep stratified sampling on our internal data, using heuristics to profile documents by layout complexity (e.g., column count, table density), linguistic rarity, and domain. To combat the inherent biases in large datasets, we purposefully over-sample from under-represented strata, such as complex multi-language tables and documents from niche scientific domains. Our 7B student model then auto-labels this high-diversity, curated pool, transforming millions of raw PDFs into a structured pre-training dataset. Ultimately, this curation-first approach ensures pre-training is not just large-scale, but high-impact—endowing the model with a deep, generalizable understanding of document structure essential for real-world parsing. 



\label{sec:target_correction}
\textbf{Stage 3: Honing the Model via Targeted Correction.}
Training on the massive corpus from the first two stages equips our base model with a broad, foundational understanding. However, this latent knowledge must be precisely honed to perform the unified parsing task with high fidelity. The objective of this final stage is therefore to align the model's emergent abilities, ensuring it can synergistically execute the core tasks of layout detection, text recognition, and relational understanding within a single forward pass.

This alignment is achieved via a systematic Human-in-the-Loop (HITL) error correction cycle, initiated by using the pre-trained model to perform inference on our diverse document pool. These initial predictions then undergo a programmatic audit, where a powerful VLM (Qwen2.5-VL-7B-Instruct~\cite{bai2025qwen2}), acting as an oracle, diagnoses a full spectrum of failure modes, identifying straightforward issues like localization inaccuracies and content or type errors, while confirming more subtle omissions and hallucinations through the examination of masked or cropped document regions.

These high-confidence, verified error cases are then routed to human annotators for efficient correction. The resulting high-signal dataset, now comprising over 15,000 samples rich in the model's specific weaknesses, provides a powerful, focused gradient that transforms the model from a broadly capable generalist into a precise and reliable parsing expert. 

\section{Experiments}
\subsection{Experimental Setting}

\paragraph{Implementation Details.} Our model, dots.ocr, consists of a 1.2B-parameter Vision Encoder (VE) and a 1.7B-parameter Language Model Decoder. It is initialized from a powerful, OCR-specialized base model (Sec. \ref{sec:model_architecture}) and then supervised finetuned (SFT) on approximately 300k diverse samples. For finetuning, we used the AdamW optimizer~\cite{loshchilov2017decoupled} with a peak learning rate of 5e-5 and a cosine decay schedule.

\paragraph{Evaluation Benchmarks and Metrics.} We conduct our primary end-to-end evaluation on OmniDocBench \cite{ouyang2025omnidocbench}, following its official protocol. We report OverallEdit, along with its component-specific scores. Supporting results on olmOCR-Bench \cite{poznanski2025olmocr} are provided in Appendix \ref{sec:appendix_olmocr_bench}, and our model's extensive multilingual capabilities are thoroughly examined in Sec. \ref{sec:multilingual_evaluation}.


\begin{table*}
\centering
\caption{State-of-the-art comparison on the \textbf{OmniDocBench} benchmark.}
\label{tab:omnidocbench_results}
\small 
\resizebox{\textwidth}{!}{%
\begin{tabular}{@{}llcccccccccccc@{}}
\toprule
\multirow{2}{*}{\textbf{Model Type}} & \multirow{2}{*}{\textbf{Methods}} & \multicolumn{2}{c}{\textbf{OverallEdit \arrowdown}} & \multicolumn{2}{c}{\textbf{TextEdit \arrowdown}} & \multicolumn{2}{c}{\textbf{FormulaEdit \arrowdown}} & \multicolumn{2}{c}{\textbf{TableTEDS \arrowup}} & \multicolumn{2}{c}{\textbf{TableEdit \arrowdown}} & \multicolumn{2}{c}{\textbf{Reading Order Edit \arrowdown}} \\
\cmidrule(lr){3-4} \cmidrule(lr){5-6} \cmidrule(lr){7-8} \cmidrule(lr){9-10} \cmidrule(lr){11-12} \cmidrule(lr){13-14} 
& & EN & ZH & EN & ZH & EN & ZH & EN & ZH & EN & ZH & EN & ZH \\
\midrule
\multirow{8}{*}{\begin{tabular}[c]{@{}l@{}}Pipeline\\ Tools\end{tabular}} 
& MinerU~\cite{wang2024mineru} & 0.150 & 0.357 & 0.061 & 0.215 & 0.278 & 0.577 & 78.6 & 62.1 & 0.180 & 0.344 &0.079 &0.292 \\
& Marker~\cite{paruchuri2024marker} & 0.336 & 0.556 & 0.080 & 0.315 & 0.530 & 0.883 & 67.6 & 49.2 & 0.619 & 0.685 & 0.114 & 0.340 \\
& Mathpix~\cite{mathpix} & 0.191 & 0.365 & 0.105 & 0.384 & 0.306 & 0.454 & 77.0 & 67.1 & 0.243 & 0.320 & 0.108 & 0.304 \\
& Docling~\cite{livathinos2025docling} & 0.589 & 0.909 & 0.416 & 0.987 & 0.999 & 1.000 & 61.3 & 25.0 & 0.627 & 0.810 & 0.313 & 0.837 \\
& Pix2Text~\cite{breezedeus2025pix2text} & 0.320 & 0.528 & 0.138 & 0.356 & 0.276 & 0.611 & 73.6 & 66.2 & 0.584 & 0.645 & 0.281 & 0.499 \\
& Unstructured~\cite{unstructured} & 0.586 & 0.716 & 0.198 & 0.481 & 0.999 & 1.000 & 0.0  & 0.06 & 1.000 & 0.998 & 0.145 & 0.387 \\
& OpenParse~\cite{filimonov2025openparse} & 0.646 & 0.814 & 0.681 & 0.974 & 0.996 & 1.000 & 64.8 & 27.5 & 0.284 & 0.639 & 0.595 & 0.641 \\
& PPStruct-V3~\cite{cui2025paddleocr} & 0.145 & 0.206 & 0.058 & 0.088 & 0.295 & 0.535 & -    & -    & 0.159 & 0.109 & 0.069 & 0.091 \\
\midrule
\multirow{9}{*}{\begin{tabular}[c]{@{}l@{}}Expert\\ VLMs\end{tabular}} 
& GOT-OCR~\cite{wei2024general} & 0.287 & 0.411 & 0.189 & 0.315 & 0.360 & 0.528 & 53.2 & 47.2 & 0.459 & 0.520 & 0.141 &0.280 \\
& Nougat~\cite{blecher2023nougat} & 0.452 & 0.973 & 0.365 & 0.998 & 0.488 & 0.941 & 39.9 & 0.0  & 0.572 & 1.000 & 0.382 & 0.954 \\
& Mistral OCR~\cite{mistral_ocr_2025} & 0.268&	0.439	&0.072	&0.325	&0.318&	0.495	&75.8&	63.6&	0.600&	0.650&	0.083&	0.284 \\
& OLMOCR-sglang~\cite{poznanski2025olmocr} & 0.326 & 0.469 & 0.097 & 0.293 & 0.455 & 0.655 & 68.1 & 61.3 & 0.608 & 0.652 & 0.145 & 0.277 \\
& SmolDocling-256M~\cite{nassar2025smoldocling} & 0.493 & 0.816 & 0.262 & 0.838 & 0.753 & 0.997 & 44.9 & 16.5 & 0.729 & 0.907 & 0.227 & 0.522 \\
& Dolphin~\cite{feng2025dolphin} & 0.206 & 0.306 & 0.107 & 0.197 & 0.447 & 0.580 & 77.3 & 67.2 & 0.180 & 0.285 &0.091 & 0.162 \\
& MinerU 2~\cite{wang2024mineru} & 0.139 & 0.240 & 0.047 & 0.109 & 0.297 & 0.536 & 82.5 & 79.0 & 0.141 & 0.195 & 0.069 & 0.118 \\
& OCRFlux~\cite{ocrflux} & 0.195 & 0.281 & 0.064 & 0.183 & 0.379 & 0.613 & 71.6 & 81.3 & 0.253 & 0.139 &0.086 & 0.187 \\
& MonkeyOCR-pro-3B~\cite{li2025monkeyocr} & 0.138 & 0.206 & 0.067 & 0.107 & \textbf{0.246} & 0.421 & 81.5 & 87.5 & 0.139 & 0.111 &0.100 &0.185 \\
\midrule
\multirow{5}{*}{\begin{tabular}[c]{@{}l@{}}General\\ VLMs\end{tabular}} 
& GPT4o~\cite{hurst2024gpt} & 0.233 & 0.399 & 0.144 & 0.409 & 0.425 & 0.606 & 72.0 & 62.9 & 0.234 & 0.329 & 0.128 & 0.251 \\
& Qwen2-VL-72B~\cite{wang2024qwen2} & 0.252 & 0.327 & 0.096 & 0.218 & 0.404 & 0.487 & 76.8 & 76.4 & 0.387 & 0.408 & 0.119 & 0.193 \\
& Qwen2.5-VL-72B~\cite{bai2025qwen2} & 0.214 & 0.261 & 0.092 & 0.180 & 0.315 & 0.434 & 82.9 & 83.9 & 0.341 & 0.262 & 0.106 & 0.168\\
& Gemini2.5-Pro~\cite{comanici2025gemini} & 0.148 & 0.212 & 0.055 & 0.168 & 0.356 & 0.439 & 85.8 & 86.4 & 0.130 & 0.119 & 0.049 & 0.121 \\
& doubao-1-5~\cite{guo2025seed1} & 0.140 & 0.162 & 0.043 & 0.085 & 0.295 & \textbf{0.384} & 83.3 & \textbf{89.3} & 0.165 & \textbf{0.085} & 0.058 & 0.094 \\
\midrule
our model & \textbf{dots.ocr} & \textbf{0.125} & \textbf{0.160} & \textbf{0.032} & \textbf{0.066} & 0.329 & 0.416 & \textbf{88.6} & 89.0 & \textbf{0.099} & 0.092 & \textbf{0.040}	& \textbf{0.067} \\
\bottomrule
\end{tabular}
}
\end{table*}

\subsection{Main Results on OmniDocBench}

Table \ref{tab:omnidocbench_results} presents a comprehensive comparison on the OmniDocBench \cite{ouyang2025omnidocbench}. The results unequivocally demonstrate the superiority of our proposed model, dots.ocr, which achieves state-of-the-art performance across nearly all key metrics in both English and Chinese.

\paragraph{Overall Performance.} dots.ocr sets a new state of the art, achieving an OverallEdit score of 0.125 in English and 0.160 in Chinese. This represents a significant improvement over the best-performing baselines, including specialized expert models like MonkeyOCR-pro-3B~\cite{li2025monkeyocr} (0.138 EN) and powerful generalist VLMs like doubao-1-5~\cite{guo2025seed1} (0.162 ZH). This holistic metric, which encapsulates all aspects of document parsing, highlights the profound benefits of our unified end-to-end learning paradigm. By jointly optimizing for detection, recognition, and structural understanding, dots.ocr effectively mitigates the error propagation inherent in pipeline-based tools and surpasses the capabilities of even the most powerful, yet fragmented, approaches.

\noindent \textbf{Component-level Dominance.} The superiority of dots.ocr is not just holistic but also evident at the component level, where it demonstrates excellence in both text recognition and structural understanding. On the one hand, it achieves a remarkable TextEdit score of 0.032 (EN) and 0.066 (ZH), significantly outperforming all other methods. This demonstrates the power of our data engine and the advantage of contextual awareness in a unified architecture. On the other hand, it excels in structural tasks, evidenced by its leading performance in TableTEDS (88.6 EN) and near-best in Chinese (89.0), as well as TableEdit. Furthermore, its state-of-the-art Reading Order Edit scores (0.040 EN, 0.067 ZH) demonstrate a superior understanding of the natural document flow, a crucial and often overlooked aspect of complex relational understanding.


In summary, the results on OmniDocBench~\cite{ouyang2025omnidocbench} robustly validate our approach. dots.ocr not only sets a new state-of-the-art but also by demonstrating the clear advantages of a truly unified, end-to-end model powered by a strategic data engine.

\subsection{Multilingual Evaluation on XDocParse}
\label{sec:multilingual_evaluation}

\textbf{XDocParse benchmark}. Existing document parsing benchmarks are predominantly composed of English and Chinese documents, failing to provide a true measure of a model's multilingual generalization capabilities. To address this critical gap, we introduce XDocParse, a new, comprehensive evaluation suite constructed from real-world documents spanning 126 languages. This benchmark serves as a challenging testbed to rigorously assess the end-to-end parsing performance of models across a vast and diverse linguistic landscape.

\begin{table*}[t]
\centering
\caption{End-to-end performance comparison on our \textbf{XDocParse} benchmark.}
\label{tab:docparse_100_results}
\resizebox{\textwidth}{!}{%
\begin{tabular}{l c c c c c c}
\toprule
\textbf{Methods} & \textbf{Overall Edit} $\downarrow$ & \textbf{Text Edit} $\downarrow$ & \textbf{Formula Edit} $\downarrow$ & \textbf{Table TEDS} $\uparrow$ & \textbf{Table Edit} $\downarrow$ & \textbf{Reading Order Edit} $\downarrow$ \\
\midrule
MonkeyOCR-3B & 0.483 & 0.445 & 0.627 & 50.93 & 0.452 & 0.409 \\
doubao-1-5-thinking-vision-pro-250428~\cite{guo2025seed1} & 0.291 & 0.226 & 0.440 & 71.20 & 0.260 & 0.238 \\
doubao-1-6~\cite{guo2025seed1} & 0.299 & 0.270 & 0.417 & 71.00 & 0.258 & 0.253 \\
Gemini-2.5-Pro~\cite{comanici2025gemini} & 0.251 & 0.163 & 0.402 & 77.10 & 0.236 & 0.202 \\
\midrule
\textbf{dots.ocr (ours)} & \textbf{0.177} & \textbf{0.075} & \textbf{0.297} & \textbf{79.20} & \textbf{0.186} & \textbf{0.152} \\ 
\bottomrule
\end{tabular}%
}
\end{table*}

\noindent \textbf{Multilingual Performance.} We compare dots.ocr with a range of strong baselines, including specialized models (MonkeyOCR-3B~\cite{li2025monkeyocr}) and powerful proprietary VLMs (doubao series~\cite{guo2025seed1}, Gemini-2.5-Pro~\cite{comanici2025gemini}). The results on our XDocParse benchmark are presented in Table~\ref{tab:docparse_100_results}.
The results unequivocally demonstrate the superiority of our approach. dots.ocr establishes a new state-of-the-art across every single metric. Notably, in the crucial Overall Edit distance, our model (0.177) achieves a 29.5\% relative reduction in error compared to the next-best model, Gemini-2.5-Pro~\cite{comanici2025gemini} (0.251). The improvement is even more dramatic in Text Edit distance, where dots.ocr (0.075) more than halves the error of Gemini-2.5-Pro~\cite{comanici2025gemini} (0.163), showing a 54\% relative reduction. This highlights the exceptional text recognition quality fostered by our unified training paradigm.


\section{Ablation study}
In this section, we conduct a series of extensive ablation studies to systematically dissect the contributions of our unified design and data-centric approach. We complement our quantitative results with qualitative examples in Appendix~\ref{sec:appendix_ablation}, which visually illustrate the key findings of our analyses.

Our experiments are designed to test three central hypotheses. First, we investigate \textbf{the synergy of joint task learning}, positing that the constituent tasks form a symbiotic triad, mutually reinforcing each other within the unified framework. Second, we aim to prove \textbf{the superiority of the unified paradigm} itself, hypothesizing that a jointly trained model develops a richer contextual representation that makes it a more powerful specialist than models trained on a single task from the outset. Finally, we examine \textbf{the efficacy of our data engine}, ablating key components of our multilingual synthesis strategy to quantify their impact on robust, cross-lingual generalization .

\subsection{The Synergy of Joint Task Learning}
\textbf{Experimental Setup.}
Our synergy ablation relies on a controlled setup built upon a training dataset comprising 116K document images. From this collection, we generate four distinct training configurations, each built upon the same set of input images but supplied with a different set of target annotations.
\begin{itemize}
    \item \textbf{M$_{\text{-Det}}$}, trained \textbf{without detection} targets to predict an ordered sequence of text.
    \item \textbf{M$_{\text{-Rec}}$}, trained \textbf{without recognition} targets to predict an ordered sequence of bounding boxes.
    \item \textbf{M$_{\text{-RO}}$}, trained \textbf{without ground-truth reading order}, relying instead on heuristic sorting schemes: horizontal (\textit{horz}), vertical (\textit{vert}), or random (\textit{rand}).
\end{itemize}

A separate model is trained for each configuration. To ensure a fair comparison, all models are trained for an identical number of epochs with the same hyperparameters and are subsequently evaluated on the comprehensive OmniDocBench \cite{ouyang2025omnidocbench}.

\noindent \textbf{Evaluation Metric.} For text recognition and reading order, we follow the official OmniDocBench \cite{ouyang2025omnidocbench} protocols. However, for layout detection, we introduce a F1-score-based metric. This is because the standard mAP metric is ill-suited for autoregressive models, as it requires confidence scores they do not natively produce and is sensitive to annotation inconsistencies. Our F1-based approach is confidence-free and more robust to inconsistencies in annotation granularity, ensuring a fairer evaluation. 
Implementation details are in Appendix \ref{sec:appendix_detection_metric}.

\noindent \textbf{Results and Analysis}. The results in Table \ref{tab:ablation_synergy} reveal a deep, synergistic interplay between the core tasks. Ablating a single task significantly impairs the performance of the remaining ones, providing clear quantitative evidence of their mutual dependence. We analyze these cross-task impacts in detail below:

\begin{table}[t]
\centering
\caption{\textbf{Ablation study on the synergy of joint task learning.} We evaluate model variants trained without specific sub-tasks to quantify their contribution.}
\label{tab:ablation_synergy}
\resizebox{\columnwidth}{!}{%
\begin{tabular}{@{}l|cccc|c@{}}
\toprule
\multirow{2}{*}{\textbf{Configuration}} & \multicolumn{2}{c}{\textbf{Overall Edit }} & \multicolumn{2}{c|}{\textbf{Reading Order Edit }} & {\textbf{Detection}} \\
\cmidrule(lr){2-3} \cmidrule(lr){4-5}
& EN & CH & EN & CH & (F1)\\
\midrule
\textbf{Unified (ours)} & \textbf{0.142} & \textbf{0.201} & \textbf{0.043} & \textbf{0.086} & 0.822 \\
\midrule 
M$_{\text{-Rec}}$ & \multicolumn{4}{c}{N/A} & \textbf{0.832} \\
M$_{\text{-Det}}$ & 0.143 & 0.211 & 0.058 & 0.108 & N/A \\
\midrule 
M$_{\text{-RO}}$ (horz) & 0.236 & 0.254 & 0.379 & 0.262 & 0.827 \\
M$_{\text{-RO}}$ (vert) & 0.214 & 0.316 & 0.315 & 0.544 & 0.825 \\
M$_{\text{-RO}}$ (rand) & 0.324 & 0.362 & 0.726 & 0.658 & 0.738 \\
\bottomrule
\end{tabular}%
}
\end{table}

\textbf{(1) Layout Detection as a Geometric Foundation.} As evidenced by our results, ablating the detection task \textbf{(M$_{\text{-Det}}$)} leads to a significant degradation in the model's ability to predict reading order. The Reading Order Edit score deteriorates by 34.9\% for English (from 0.043 to 0.058) and 25.6\% for Chinese (from 0.086 to 0.108), accompanied by a slight increase in Overall Edit. This confirms that a precise spatial detection is a prerequisite for inferring the document's logical flow. Without it, the model struggles to reconstruct a coherent text sequence from disembodied content.

\textbf{(2) Text Recognition as a Semantic Regularizer.} Text recognition imposes a powerful semantic constraint on the detection task. Counter-intuitively, ablating recognition \textbf{(M$_{\text{-Rec}}$)} results in a slight improvement in the detection score (0.832 vs. 0.822). We posit this is not because recognition is irrelevant, but because it acts as a beneficial regularizer. In the unified model, the detection objective is not merely to find any box, but to find boxes conducive to recognition. This semantic-driven pressure guides the detector toward more legible text regions, ultimately benefiting the end-to-end goal, even at the cost of a marginal drop in the isolated detection metric.

\textbf{(3) Reading Order as a Structural-Semantic Bridge.} A coherent reading order signal is critical for bridging low-level vision with high-level structure. This is evident as we degrade the training sequence from heuristic sorting (vertical, horizontal) to random shuffling. Not only do the text and order metrics worsen, but the detection score also plummets from 0.827 to 0.738. This demonstrates that an incoherent sequence does more than just confuse the final output, it corrupts the model's internal representation of the document layout. It suggests that in VLMs, fundamental visual perception is not learned in isolation but is deeply intertwined with structural and semantic understanding.

In summary, these ablation studies reveal that layout detection, text recognition, and relational understanding are not independent modules but a symbiotic triad. Within our unified VLM, each task provides crucial inductive biases that guide and regularize the others. This reciprocal reinforcement leads to a level of holistic and robust document comprehension that fragmented, multi-stage pipelines cannot achieve. 

\subsection{Superiority of the Unified Paradigm}
\textbf{Experimental Setup.}
Having established the synergy between tasks, we now aim to prove the inherent superiority of our unified paradigm itself. We hypothesize that the rich contextual understanding gained during joint training makes our model a more powerful \textbf{specialist}—outperforming models trained solely on a single task from the outset. To isolate this effect, we design an experiment to disentangle the benefits of unified training from unified inference. We compare three distinct configurations:
    
    

\begin{itemize}
    \item \textbf{U $\rightarrow$ U (Unified $\rightarrow$ Unified)}. Our full dots.ocr model, trained and evaluated on all tasks jointly. This represents our full end-to-end paradigm.
    
    \item \textbf{U $\rightarrow$ S (Unified $\rightarrow$ Specialist)}. The same jointly-trained model, but evaluated on a single, specialized task at inference time. This isolates the benefit of unified training.
    
    \item \textbf{S $\rightarrow$ S (Specialist $\rightarrow$ Specialist)}. A baseline model trained and evaluated on only a single, specialized task. This configuration represents the dominant paradigm of existing specialized methods.
\end{itemize}
%
\begin{table}[t]
\centering
\caption{\textbf{Superiority of the Unified Paradigm.} We disentangle the effects of unified training from unified inference by comparing our full model (\textbf{U~$\rightarrow$~U}) against specialist configurations (\textbf{U~$\rightarrow$~S} and \textbf{S~$\rightarrow$~S}).}
\label{tab:ablation_paradigm}
\resizebox{\columnwidth}{!}{%
\begin{tabular}{@{}c|cccc|c@{}}
\toprule
\multirow{2}{*}{\textbf{Configuration}} & \multicolumn{2}{c}{\textbf{Overall Edit }} & \multicolumn{2}{c|}{\textbf{Reading Order Edit }} & \textbf{Detection} \\
\cmidrule(lr){2-3} \cmidrule(lr){4-5}
& EN & CH & EN & CH & (F1)\\
\midrule
\textbf{U $\rightarrow$ U} & \textbf{0.143} & \textbf{0.187} & \textbf{0.045} & \textbf{0.086} & 0.820 \\
\textbf{U $\rightarrow$ S} & 0.148 & 0.204 & 0.055 & 0.103 & 0.829 \\
\textbf{S $\rightarrow$ S} & \textbf{0.143} & 0.211 & 0.058 & 0.108 & \textbf{0.832} \\ \bottomrule
\end{tabular}%
}
\end{table}
\noindent \textbf{Results and Analysis}. Table \ref{tab:ablation_paradigm} dissects the advantages of our unified paradigm by comparing our full model (\textbf{U $\rightarrow$ U}), its specialist counterpart (\textbf{U $\rightarrow$ S}), and a purely specialized baseline (\textbf{S $\rightarrow$ S}), revealing two distinct patterns:

\textbf{(1) Superiority in Context-Rich Tasks.} For text recognition and reading order, a clear hierarchy emerges: \textbf{U $\rightarrow$ U $>$ U $\rightarrow$ S $>$ S $\rightarrow$ S}. The crucial comparison is between \textbf{U $\rightarrow$ S} and \textbf{S $\rightarrow$ S}. The jointly-trained \textbf{U $\rightarrow$ S} model consistently outperforms the \textbf{S $\rightarrow$ S} baseline (e.g., Overall Edit on Chinese drops from 0.211 to 0.204), providing direct evidence that unified training alone imbues the model with superior contextual awareness. The fully unified \textbf{U $\rightarrow$ U} model further extends this lead, confirming that both joint training and joint inference contribute to peak performance.


\textbf{(2) Stability in Pure Localization.} In contrast, for the pure detection task, performance is remarkably stable across all three configurations. The specialized \textbf{S $\rightarrow$ S} model achieves a marginally higher detection score (0.832 vs. 0.829 for \textbf{U $\rightarrow$ S}). This minimal difference suggests that while specialized training can narrowly optimize for localization, this task is not meaningfully compromised within our richer, joint-training framework.

In aggregate, these results show that while a specialized model holds a negligible edge in detection, our unified paradigm yields substantial gains in the more complex tasks of recognition and relational understanding. This superiority stems from the richer contextual representation forged through joint learning—a prerequisite for true document comprehension.
\subsection{Effectiveness of the Holistic Data Engine}
To validate the data engine's effectiveness, we conduct a systematic ablation study. For this purpose, We focus on three critical types of data generated by our engine, each targeting a key challenge in document understanding:
\begin{itemize}
    \item \textbf{D-Multilingual}: The full dataset excluding the Unified Multilingual Data. This variant is trained without synthetic documents spanning 100+ languages, allowing us to measure the impact on multilingual generalization.
    
    \item \textbf{D-Structured}: The full dataset excluding the Structured Data. This ablates the specialized set rich in complex tables and formulas, testing the model's fine-grained parsing capabilities.
    
    \item \textbf{D-Correction}: Isolates the impact of the targeted correction data on overall performance
\end{itemize}


\begin{table}[t]
\centering
\caption{\textbf{Ablation study on our data engine.} We quantify the impact of each data pillar by removing it from the full training configuration.}
\label{tab:data_ablation}
\resizebox{\columnwidth}{!}{%
\begin{tabular}{@{}l|cc|cc|c@{}}
\toprule
\multirow{2}{*}{\textbf{Configuration}} & \multicolumn{2}{c}{\textbf{Overall Edit}} & \multicolumn{2}{c|}{\textbf{Reading Order}} & \textbf{Detection} \\
\cmidrule(lr){2-3} \cmidrule(lr){4-5}
& EN & CH & EN & CH & (F1) \\
\midrule
\textbf{dots.ocr (Full)} & \textbf{0.125} & \textbf{0.160} & \textbf{0.040} & \textbf{0.067} & \textbf{0.849} \\
D-Multilingual & 0.135 & 0.171 & 0.051 & 0.083 & 0.811 \\
D-Structured & 0.145 & 0.181 & 0.047 & 0.073 & 0.820 \\
D-Correction & 0.136 & 0.183 & 0.045 & 0.093 & 0.788 \\
\bottomrule
\end{tabular}
}
\end{table}

The results in Table~\ref{tab:data_ablation} reveal a clear hierarchy of impact among our data pillars, while simultaneously underscoring that each is indispensable.

The most dramatic impact comes from ablating the targeted correction data \textbf{(w/o D-Correction)}. This single change causes a catastrophic drop in detection performance, with the detection score plummeting from 0.849 to 0.788. This provides unequivocal evidence that our target correction loop is the critical mechanism for patching the model's perceptual weaknesses and achieving high-fidelity localization.

While not as dramatic, the removal of the other two pillars reveals their distinct and vital roles. Ablating the structured data \textbf{(w/o D-Structured)} disproportionately harms the model's ability to parse complex Chinese documents (Overall Edit degrades from 0.160 to 0.181), confirming this data's crucial role in teaching structural understanding. Similarly, removing the multilingual data \textbf{(w/o D-Multilingual)} leads to a broad degradation across all metrics, highlighting its foundational importance for building a robust, cross-lingually capable model.

In summary, these ablations demonstrate that our model's performance stems not from any single data source, but from the synergistic effect of our holistic data engine. Its components play distinct, complementary roles to forge a comprehensively robust model, ensuring high fidelity across a vast spectrum of languages, layouts, and complexities.

\section{Conclusion}
In this work, we presented dots.ocr, a unified model powered by a holistic data engine. Its architecture and data engine have enabled it to establish state-of-the-art performance across extensive benchmarks. Beyond its immediate performance, we argue that the true significance of this work lies in its potential to fuel the next generation of Vision-Language Models. By holistically parsing vast document corpora, dots.ocr can serve as a powerful data engine, unlocking novel pre-training tasks on rich, structured, and grounded data. We believe this is a pivotal step towards teaching VLMs to not merely see pixels, but to truly comprehend the world’s structured knowledge. 
As we posited in the main paper's conclusion, the ultimate value of dots.ocr extends beyond its state-of-the-art performance on document analysis benchmarks. We argued that its true significance lies in its potential to serve as a powerful data engine to fuel the next generation of Vision-Language Models (VLMs). In this section, we provide a detailed elaboration of this vision.

{
    \small
    \bibliographystyle{ieeenat_fullname}
    \bibliography{main}
}

\clearpage
\onecolumn
\maketitlesupplementary


\section{Supporting Results on olmOCR-Bench}
\label{sec:appendix_olmocr_bench}

In this section, we further verify the performance of our model by presenting supplementary end-to-end evaluation results on the olmOCR-Bench benchmark~\cite{poznanski2025olmocr}, complementing the main findings reported in the paper. Detailed results of dots.ocr and various baselines are shown in Table~\ref{tab:olm_breakdown}.

\begin{table*}[ht]
\centering
\caption{Detailed comparison on the \textbf{olmOCR-Bench \cite{poznanski2025olmocr}} benchmark. All scores are reported as a percentage (\%), where higher is better.}
\label{tab:olm_breakdown}
\resizebox{\textwidth}{!}{%
\begin{tabular}{@{}lccccccccc@{}}
\toprule
\textbf{Model} & \textbf{ArXiv} & \textbf{Old Scans Math} & \textbf{Tables} & \textbf{Old Scans} & \textbf{\begin{tabular}[c]{@{}c@{}}Headers and\\ Footers \end{tabular}} & \textbf{\begin{tabular}[c]{@{}c@{}}Multi\\ column \end{tabular}} & \textbf{\begin{tabular}[c]{@{}c@{}}Long Tiny\\ Text \end{tabular}} & \textbf{Base} & \textbf{Overall} \\
\midrule
GOT-OCR~\cite{wei2024general} & 52.7 & 52.0 & 0.2 & 22.1 & 93.6 & 42.0 & 29.9 & 94.0 & 48.3 $\pm$ 1.1 \\
Marker~\cite{paruchuri2024marker} & 76.0 & 57.9 & 57.6 & 27.8 & 84.9 & 72.9 & 84.6 & 99.1 & 70.1 $\pm$ 1.1 \\
MinerU~\cite{wang2024mineru} & 75.4 & 47.4 & 60.9 & 17.3 & \textbf{96.6} & 59.0 & 39.1 & 96.6 & 61.5 $\pm$ 1.1 \\
Mistral OCR~\cite{mistral_ocr_2025} & 77.2 & 67.5 & 60.6 & 29.3 & 93.6 & 71.3 & 77.1 & 99.4 & 72.0 $\pm$ 1.1 \\
Nanonets OCR~\cite{docext} & 67.0 & 68.6 & 77.7 & 39.5 & 40.7 & 69.9 & 53.4 & 99.3 & 64.5 $\pm$ 1.1 \\
GPT4o (No Anchor)~\cite{hurst2024gpt} & 51.5 & \textbf{75.5} & 69.1 & 40.9 & 94.2 & 68.9 & 54.1 & 96.7 & 68.9 $\pm$ 1.1 \\
GPT4o (Anchored)~\cite{hurst2024gpt} & 53.5 & 74.5 & 70.0 & 40.7 & 93.8 & 69.3 & 60.6 & 96.8 & 69.9 $\pm$ 1.1 \\
Gemini Flash 2 (No Anchor)~\cite{comanici2025gemini} & 32.1 & 56.3 & 61.4 & 27.8 & 48.0 & 58.7 & \textbf{84.4} & 94.0 & 57.8 $\pm$ 1.1 \\
Gemini Flash 2 (Anchored)~\cite{comanici2025gemini} & 54.5 & 56.1 & 72.1 & 34.2 & 64.7 & 61.5 & 71.5 & 95.6 & 63.8 $\pm$ 1.2 \\
Qwen2-VL-72B (No Anchor)~\cite{wang2024qwen2} & 19.7 & 31.7 & 24.2 & 17.1 & 88.9 & 8.3 & 6.8 & 55.5 & 31.5 $\pm$ 0.9 \\
 Qwen2.5-VL-72B (No Anchor)~\cite{bai2025qwen2} & 63.1 & 65.7 & 67.3 & 38.6 & 73.6 & 68.3 & 49.1 & 98.3 & 65.5 $\pm$ 1.2 \\
olmOCR v0.1.75 (No Anchor)~\cite{poznanski2025olmocr} & 71.5 & 71.4 & 71.4 & \textbf{42.8} & 94.1 & 77.7 & 71.0 & 97.8 & 74.7 $\pm$ 1.1 \\
olmOCR v0.1.75 (Anchored)~\cite{poznanski2025olmocr} & 74.9 & 71.2 & 71.0 & 42.2 & 94.5 & 78.3 & 73.3 & 98.3 & 75.5 $\pm$ 1.0 \\
MonkeyOCR-pro-3B~\cite{li2025monkeyocr} & \textbf{83.8} & 68.8 & 74.6 & 36.1 & 91.2 & 76.6 & 80.1 & 95.3 & 75.8 $\pm$ 1.0 \\
\textbf{\texttt{dots.ocr} (Ours)} & 82.1 & 64.2 & \textbf{88.3} & 40.9 & 94.1 & \textbf{82.4} & 81.2 & \textbf{99.5} & \textbf{79.1 $\pm$ 1.0} \\
\bottomrule
\end{tabular}%
}
\end{table*}



\section{Layout Detection Metric}
\label{sec:appendix_detection_metric}


In this section, We introduce a confidence-free, F1-score-based evaluation metric that is more robust and provides a fairer assessment of layout detection performance.

\textbf{Core Idea.} The core idea is a two-stage matching process. The first stage handles clear one-to-one matches between predicted and ground-truth boxes. The second stage addresses more complex scenarios, such as one-to-many or many-to-many relationships, by first clustering spatially related boxes into larger ``super-boxes" and then attempting to match them. This approach allows the metric to correctly reward a model that, for example, predicts three separate lines for a paragraph that is annotated as a single block in the ground truth.

\textbf{Category-Aware and Category-Agnostic Modes.}
Our evaluation metric supports two modes. In the \emph{category-aware} mode, all operations (matching and clustering) are carried out within each annotation category, only boxes of the same category are matched together. The True Positives (TPs), False Positives (FPs), and False Negatives (FNs) are counted for each category, and then aggregated. In the \emph{category-agnostic} mode, the process ignores box categories, treating all boxes as a single group. This flexibility allows evaluation of either pure layout prediction quality or joint layout and category performance.

\textbf{Two-Stage Matching Process.}
The evaluation logic is detailed in Algorithm \ref{alg:f1_metric}. It proceeds as follows:

\textbf{Stage 1: One-to-One Matching.} We first perform an optimal bipartite matching between the set of all predicted boxes ($\mathcal{P}$) and all ground-truth boxes ($\mathcal{G}$) using the Hungarian algorithm on an IoU-based cost matrix. Matched pairs with an IoU greater than a predefined threshold (e.g., 0.5) are counted as initial TPs. These matched boxes are then removed from their respective sets, leaving unmatched predicted boxes ($\mathcal{P}'$) and unmatched ground-truth boxes ($\mathcal{G}'$).

\textbf{Stage 2: Matching via Clustering.} The remaining unmatched boxes, $\mathcal{P}'$ and $\mathcal{G}'$, are processed independently. We apply a clustering algorithm to each set. This algorithm groups spatially adjacent boxes based on geometric heuristics (e.g., high vertical overlap and small horizontal gap for line merging, followed by high horizontal overlap and small vertical gap for paragraph merging). This step results in a set of merged predicted boxes ($\mathcal{P}_{merged}$) and merged ground-truth boxes ($\mathcal{G}_{merged}$). We then perform a second round of optimal bipartite matching on these merged sets. Pairs with an IoU above the threshold contribute additional TPs.

\textbf{Final Calculation.} After the second stage, any remaining unmatched boxes in $\mathcal{P}_{merged}$ are counted as FPs, and any remaining boxes in $\mathcal{G}_{merged}$ are counted as FNs. The total TPs, FPs, and FNs are then used to calculate the final Precision, Recall, and F1-score.


\begin{algorithm*}[ht]
\DontPrintSemicolon
\SetAlgoLined
\KwInput{Predicted boxes $\mathcal{P}$, Ground-truth boxes $\mathcal{G}$, IoU threshold $\tau$}
\KwOutput{Precision, Recall, F1-score}
\BlankLine

\tcp{Stage 1: One-to-one matching}
$M_1 \leftarrow \text{OptimalBipartiteMatch}(\mathcal{P}, \mathcal{G})$ \tcp*{Using Hungarian algorithm on IoU}
$TP_1 \leftarrow 0$\;
$\mathcal{P}_{matched} \leftarrow \emptyset$, $\mathcal{G}_{matched} \leftarrow \emptyset$\;
\ForEach{match $(p, g) \in M_1$}{
    \If{$\text{IoU}(p, g) \geq \tau$}{
        $TP_1 \leftarrow TP_1 + 1$\;
        $\mathcal{P}_{matched} \leftarrow \mathcal{P}_{matched} \cup \{p\}$\;
        $\mathcal{G}_{matched} \leftarrow \mathcal{G}_{matched} \cup \{g\}$\;
    }
}
$\mathcal{P}' \leftarrow \mathcal{P} \setminus \mathcal{P}_{matched}$ \tcp*{Unmatched predicted boxes}
$\mathcal{G}' \leftarrow \mathcal{G} \setminus \mathcal{G}_{matched}$ \tcp*{Unmatched ground-truth boxes}
\BlankLine

\tcp{Stage 2: Matching after clustering}
$\mathcal{P}_{merged} \leftarrow \text{ClusterBoxes}(\mathcal{P}')$ \tcp*{Merge adjacent boxes in $\mathcal{P}'$}
$\mathcal{G}_{merged} \leftarrow \text{ClusterBoxes}(\mathcal{G}')$ \tcp*{Merge adjacent boxes in $\mathcal{G}'$}
\BlankLine

$M_2 \leftarrow \text{OptimalBipartiteMatch}(\mathcal{P}_{merged}, \mathcal{G}_{merged})$\;
$TP_2 \leftarrow 0$\;
$\mathcal{P}_{merged\_matched} \leftarrow \emptyset$, $\mathcal{G}_{merged\_matched} \leftarrow \emptyset$\;
\ForEach{match $(p_m, g_m) \in M_2$}{
    \If{$\text{IoU}(p_m, g_m) \geq \tau$}{
        $TP_2 \leftarrow TP_2 + 1$\;
        $\mathcal{P}_{merged\_matched} \leftarrow \mathcal{P}_{merged\_matched} \cup \{p_m\}$\;
        $\mathcal{G}_{merged\_matched} \leftarrow \mathcal{G}_{merged\_matched} \cup \{g_m\}$\;
    }
}
\BlankLine

\tcp{Final Calculation}
$TP_{total} \leftarrow TP_1 + TP_2$\;
$FP \leftarrow |\mathcal{P}_{merged}| - |\mathcal{P}_{merged\_matched}|$\;
$FN \leftarrow |\mathcal{G}_{merged}| - |\mathcal{G}_{merged\_matched}|$\;
\BlankLine
$Precision \leftarrow TP_{total} / (TP_{total} + FP)$\;
$Recall \leftarrow TP_{total} / (TP_{total} + FN)$\;
$F1 \leftarrow 2 \times (Precision \times Recall) / (Precision + Recall)$\;
\BlankLine
\Return $Precision, Recall, F1$\;
\caption{Layout Detection Metric}
\label{alg:f1_metric}
\end{algorithm*}

\section{Qualitative Analysis of Ablation Studies}
\label{sec:appendix_ablation}
In this section, we provide qualitative examples that visually complement the quantitative results presented in our ablation study. These examples are chosen to vividly illustrate the core findings of our investigation, particularly the nuanced roles of text recognition and reading order within our unified framework.

\begin{figure}[ht]
\centering

\begin{minipage}{0.32\linewidth}
\centering
\textbf{(a) Ground Truth} \\[4pt]
\fbox{\includegraphics[width=0.9\linewidth]{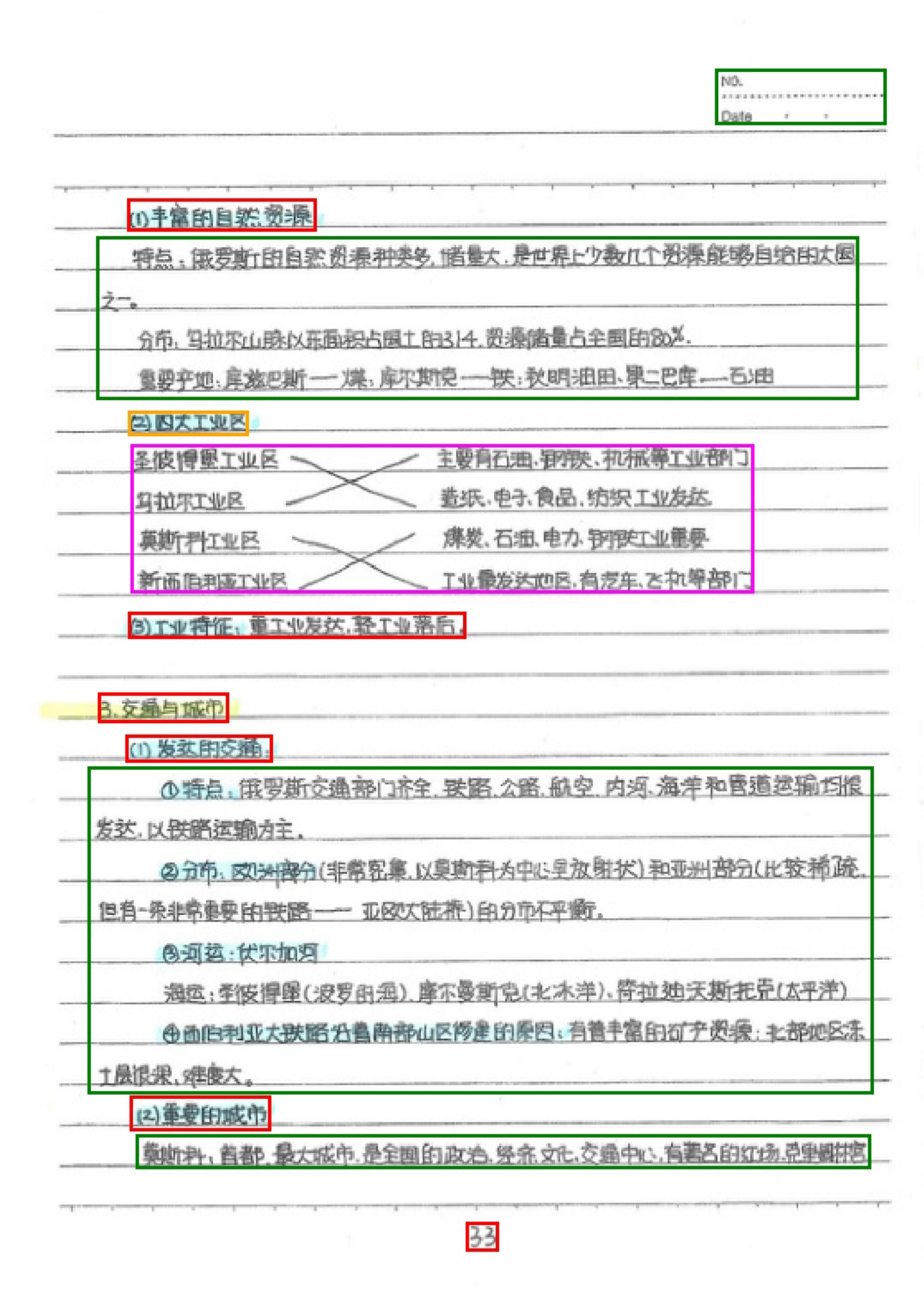}}
\end{minipage}
\hfill
\begin{minipage}{0.32\linewidth}
\centering
\textbf{(b) Our Unified Model} \\[4pt]
\fbox{\includegraphics[width=0.9\linewidth]{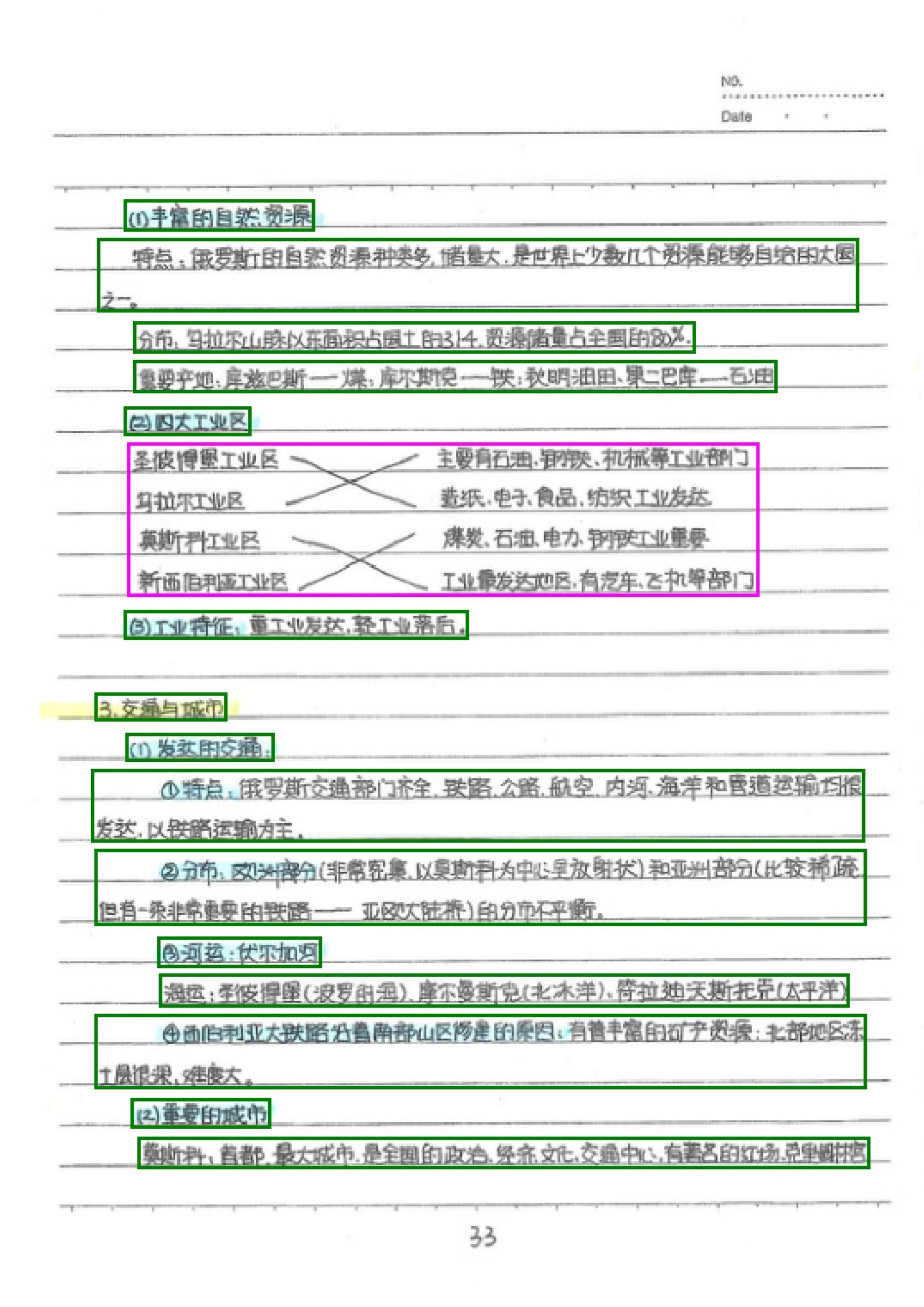}}
\end{minipage}
\hfill
\begin{minipage}{0.32\linewidth}
\centering
\textbf{(c) M$_{\text{-Rec}}$ (without recognition)} \\[4pt]
\fbox{\includegraphics[width=0.9\linewidth]{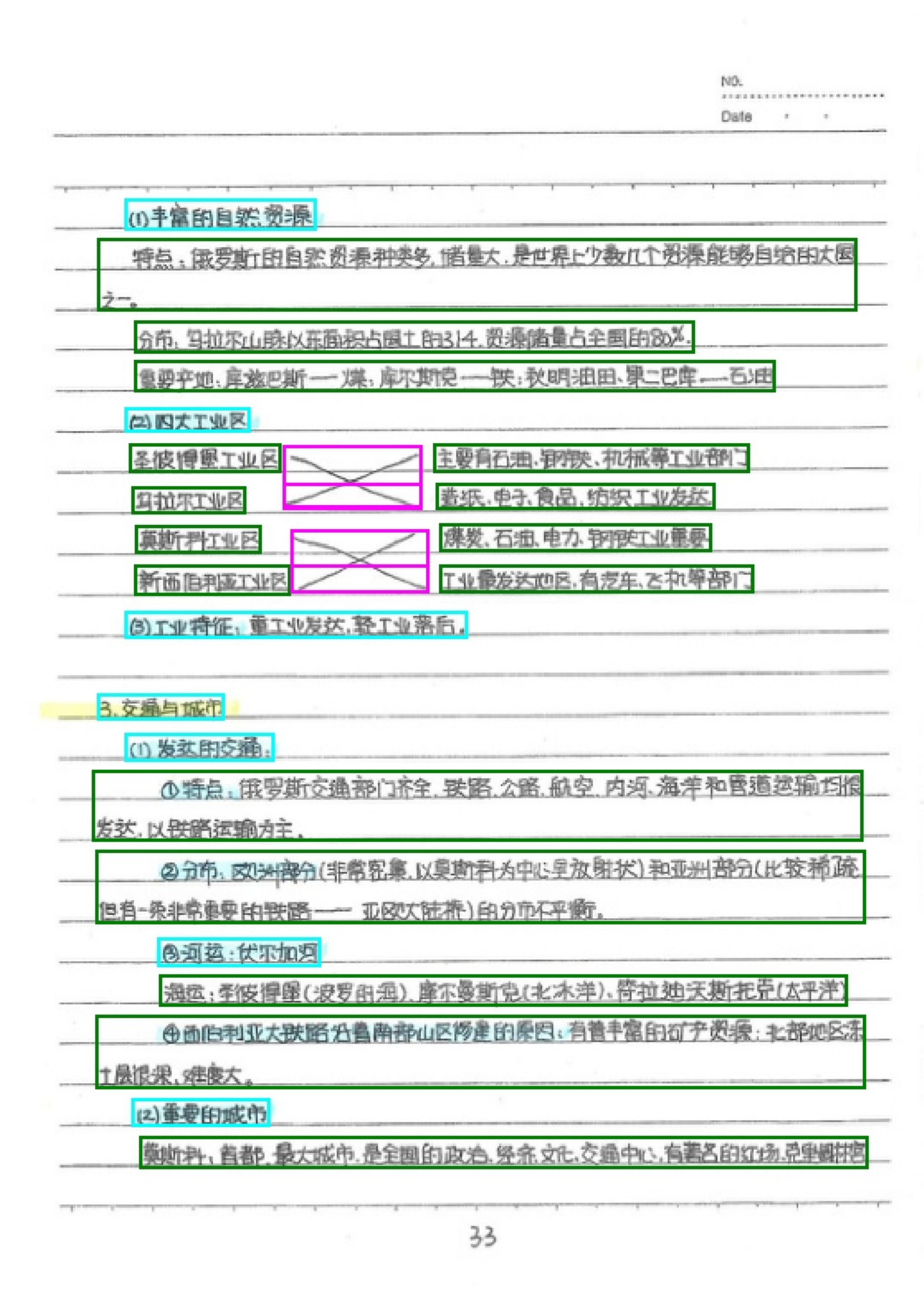}}
\end{minipage}

\caption{
\textbf{Effect of Removing the Recognition task (1).}
(a) Ground Truth. 
(b) Output of our model under the unified training paradigm, which correctly preserves the entire connected diagram as a single coherent structure. 
(c) Output of the model without the recognition task. 
Unlike (b), the ablated model breaks the originally unified connected diagram into many isolated fragments, treating each short stroke or text segment as an individual object. This demonstrates that the recognition task provides essential semantic constraints that prevent such structural fragmentation.
}
\label{fig:ablation_recognition1}
\end{figure}

\begin{figure}[ht]
\centering

\begin{minipage}{0.32\linewidth}
\centering
\textbf{(a) Ground Truth} \\[4pt]
\fbox{\includegraphics[width=0.9\linewidth]{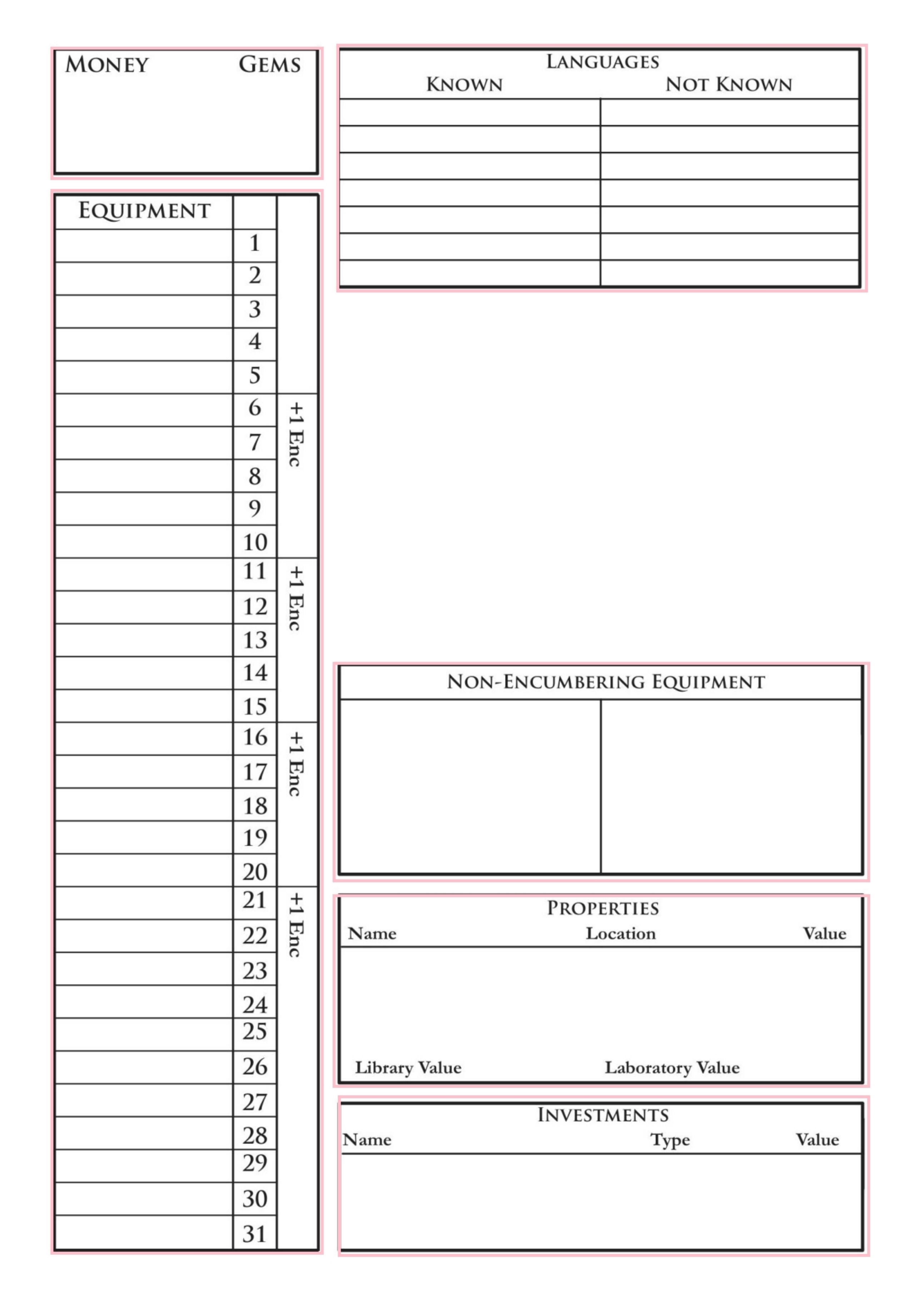}}
\end{minipage}
\hfill
\begin{minipage}{0.32\linewidth}
\centering
\textbf{(b) Our Unified Model} \\[4pt]
\fbox{\includegraphics[width=0.9\linewidth]{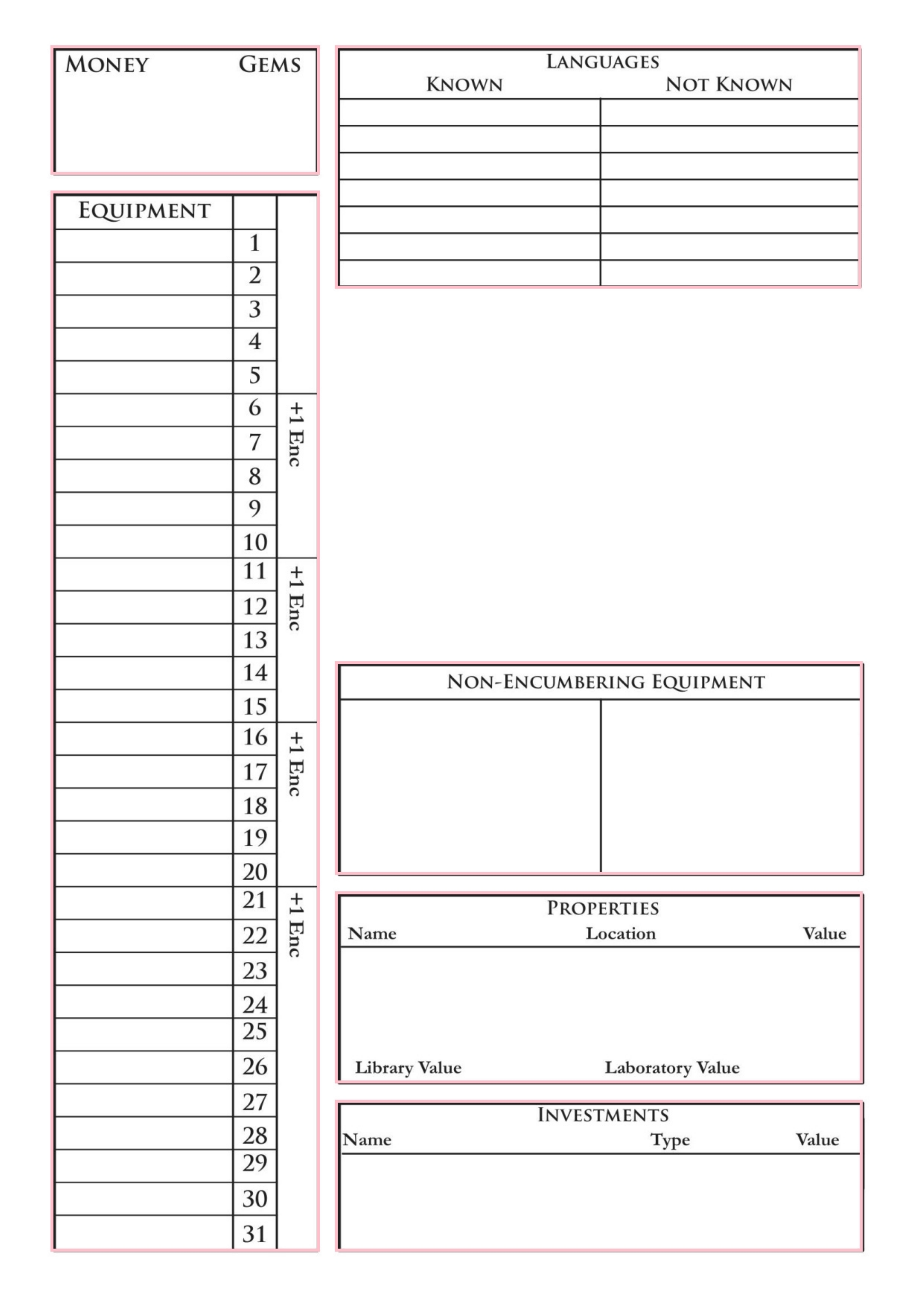}}
\end{minipage}
\hfill
\begin{minipage}{0.32\linewidth}
\centering
\textbf{(c) M$_{\text{-Rec}}$ (without recognition)} \\[4pt]
\fbox{\includegraphics[width=0.9\linewidth]{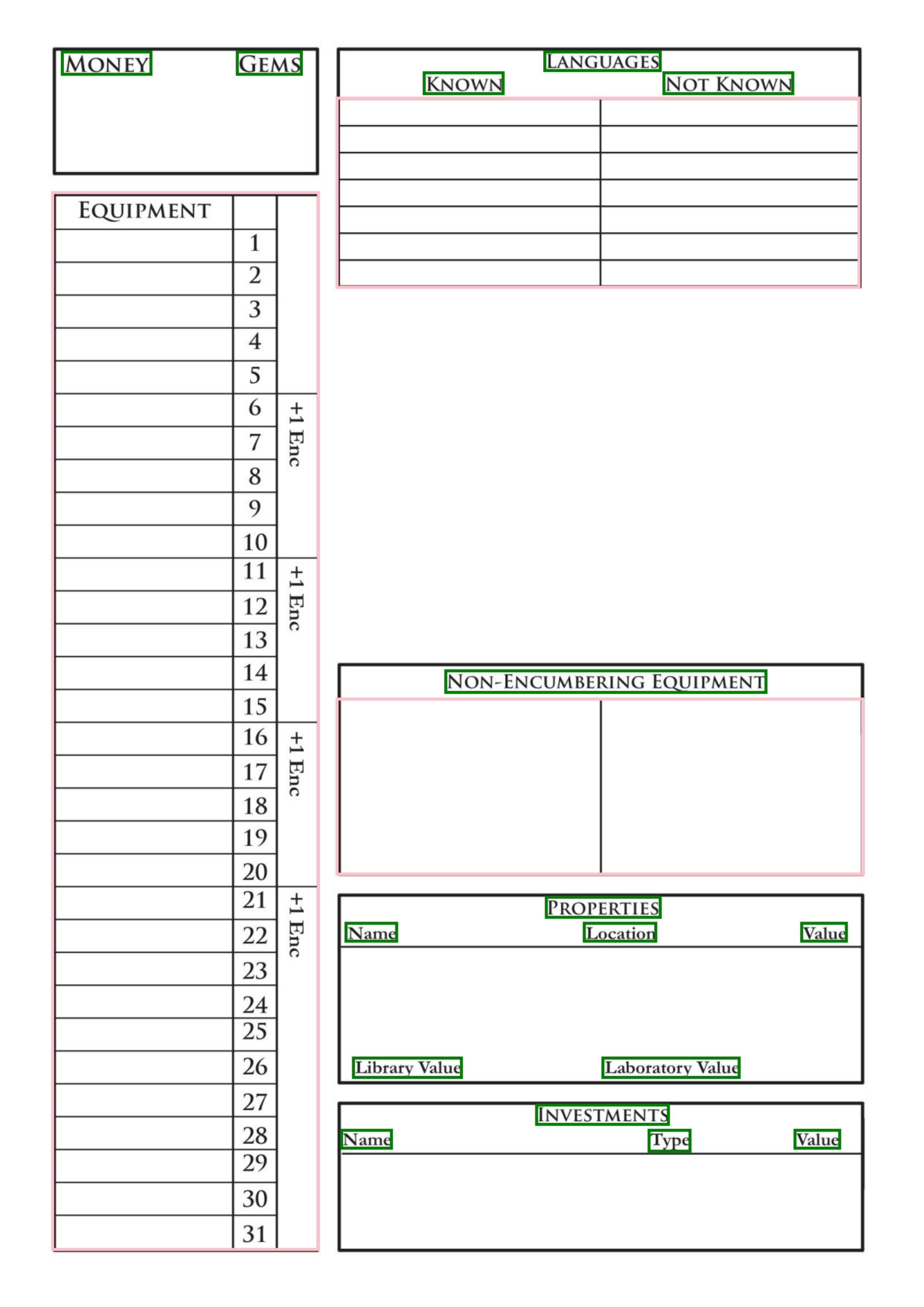}}
\end{minipage}

\caption{
\textbf{Effect of Removing the Recognition task (2).}
For the table in the lower-right region,
(b) Our unified model correctly recognizes the entire table as a single structured layout, preserving its rows, columns, and bounding geometry. 
(c) In contrast, the model without the recognition objective fails to detect the table as a unified structure. 
Instead of capturing the table boundaries and grid lines, it isolates and boxes only the individual text tokens inside the table. 
}
\label{fig:ablation_recognition2}
\end{figure}

\begin{figure}[ht]
\centering

\begin{minipage}{0.32\linewidth}
\centering
\textbf{(a) Ground Truth} \\[4pt]
\fbox{\includegraphics[width=0.9\linewidth]{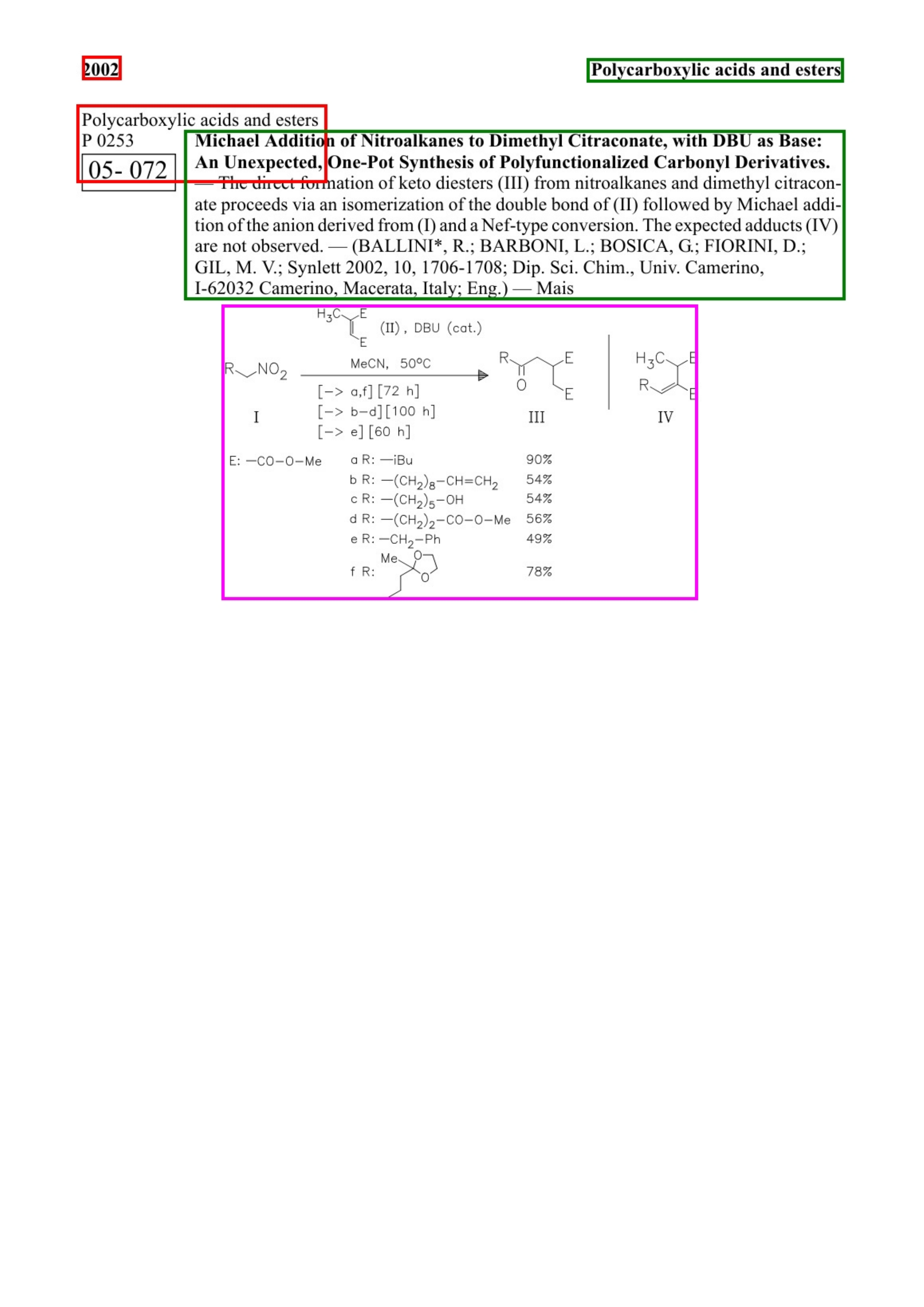}}
\end{minipage}
\hfill
\begin{minipage}{0.32\linewidth}
\centering
\textbf{(b) Our Unified Model} \\[4pt]
\fbox{\includegraphics[width=0.9\linewidth]{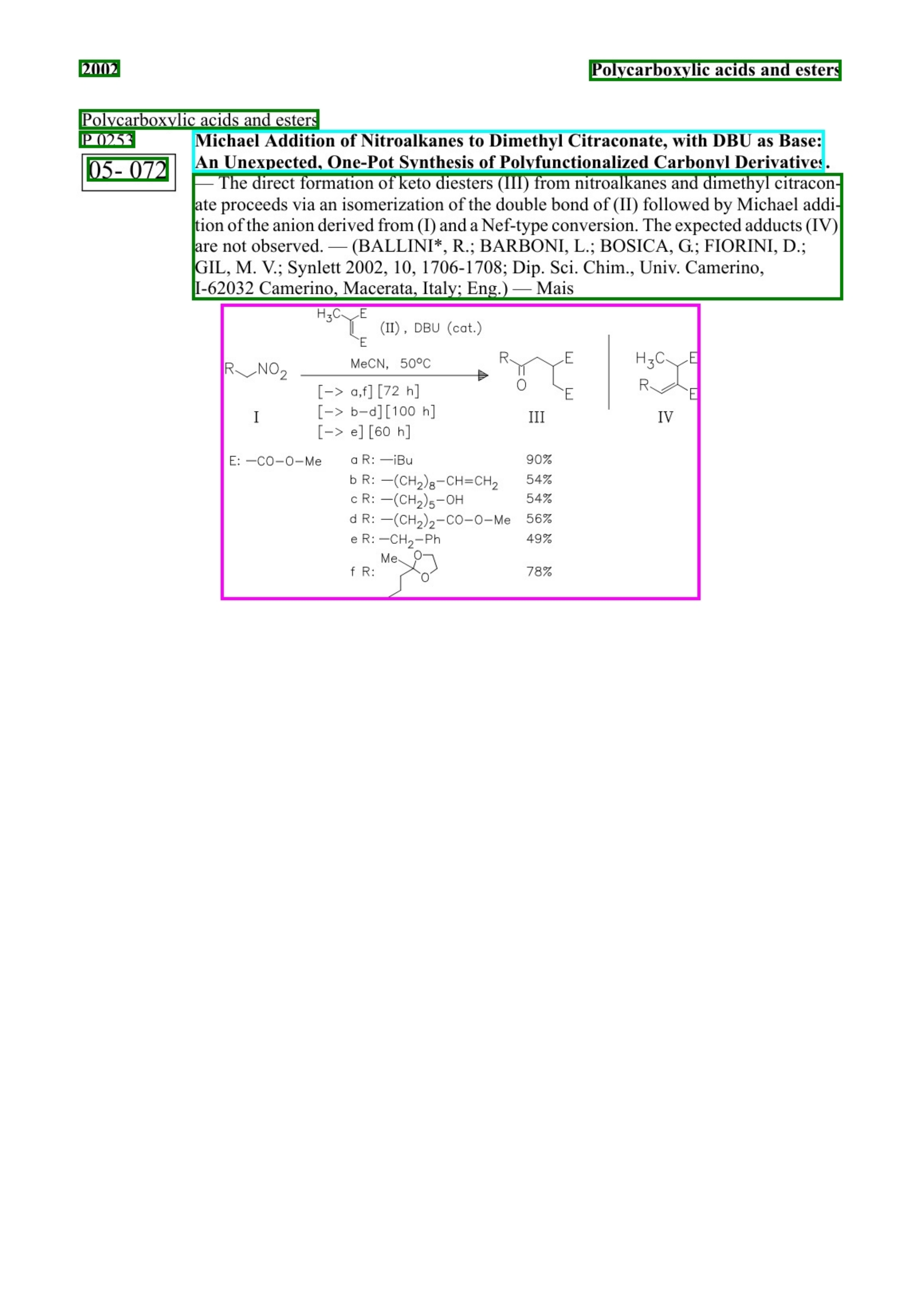}}
\end{minipage}
\hfill
\begin{minipage}{0.32\linewidth}
\centering
\textbf{(c) M$_{\text{-Rec}}$ (without recognition)} \\[4pt]
\fbox{\includegraphics[width=0.9\linewidth]{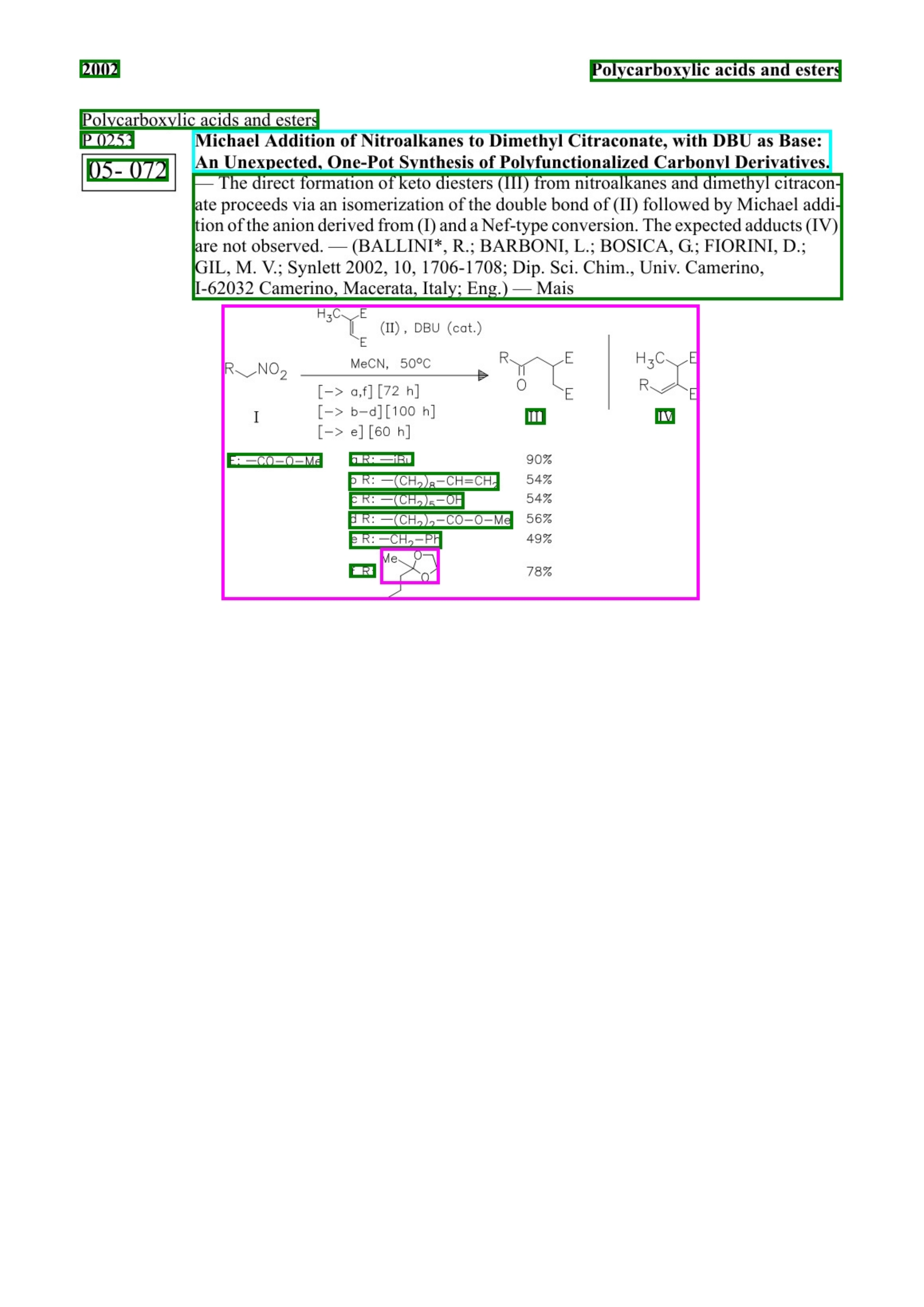}}
\end{minipage}

\caption{
\textbf{Effect of Removing the Recognition task (3).}
For the chemical reaction scheme in the lower region,
(b) Our unified model detects the entire reaction diagram as a single, coherent figure, preserving its hierarchical structure where internal elements (chemical formulas, labels, yields, arrows) are treated as components of one unified visual unit.
(c) Although the recognition-ablated model successfully delineates the overall diagram, it disrupts the internal hierarchy through excessive segmentation. By individually isolating text tokens, chemical formulas, and captions, the model produces a cluttered and structurally inconsistent interpretation.
}
\label{fig:ablation_recognition3}
\end{figure}

\begin{figure}[ht]
\centering

\begin{minipage}{0.32\linewidth}
\centering
\textbf{(a) Ground Truth} \\[4pt]
\fbox{\includegraphics[width=0.9\linewidth]{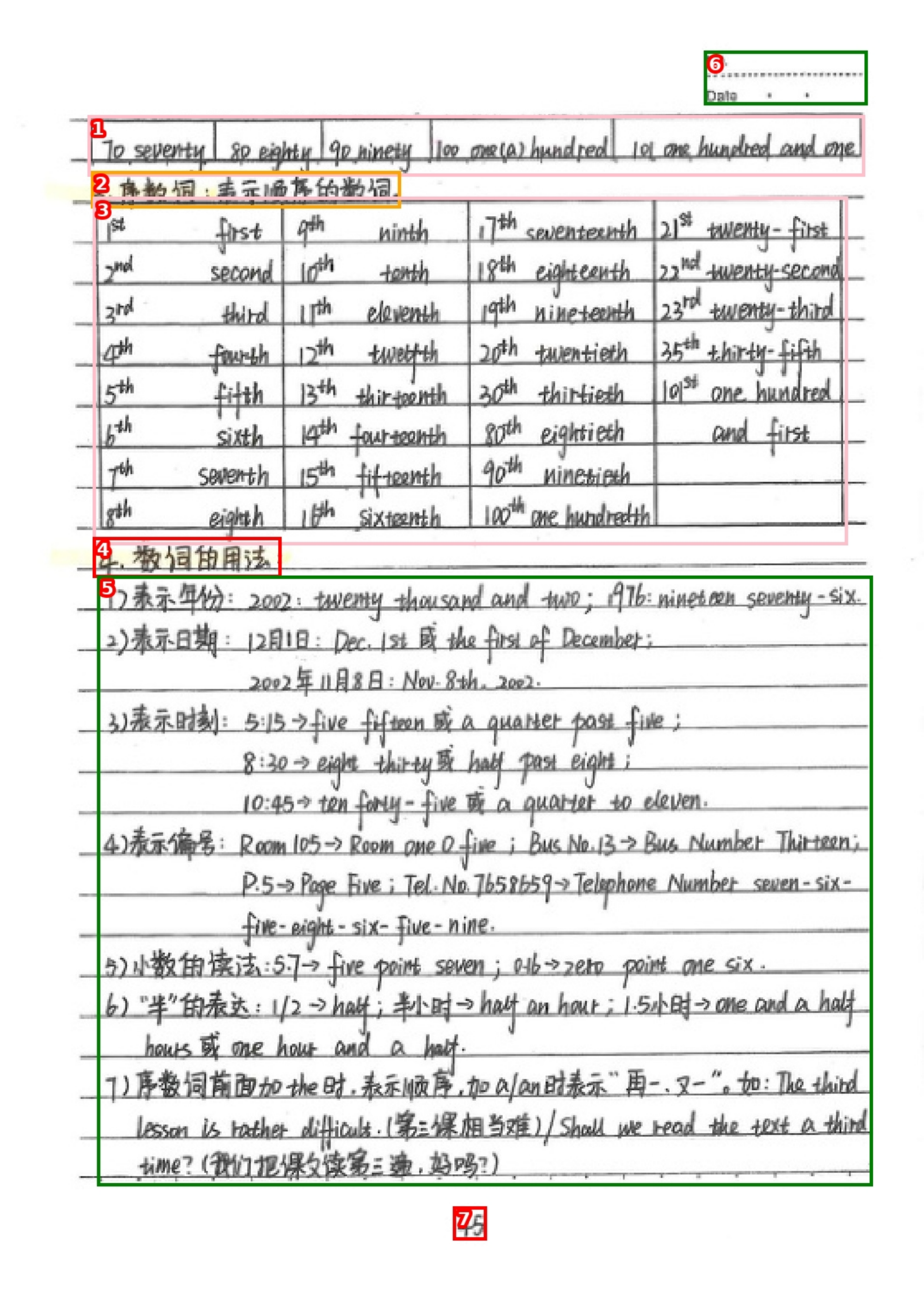}}
\end{minipage}
\hfill
\begin{minipage}{0.32\linewidth}
\centering
\textbf{(b) Our Unified Model} \\[4pt]
\fbox{\includegraphics[width=0.9\linewidth]{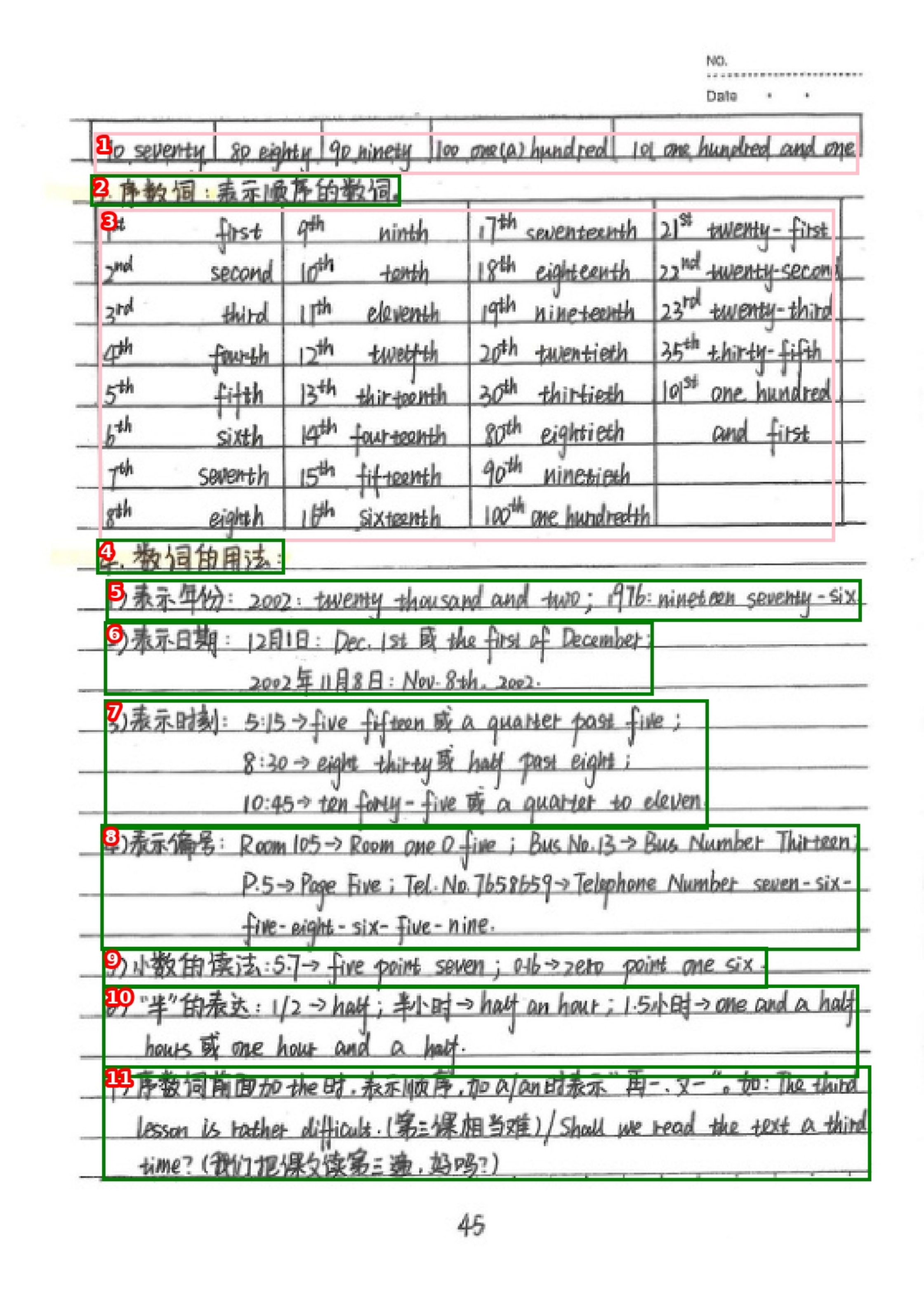}}
\end{minipage}
\hfill
\begin{minipage}{0.32\linewidth}
\centering
\textbf{(c) M$_{\text{-Rec}}$ (without recognition)} \\[4pt]
\fbox{\includegraphics[width=0.9\linewidth]{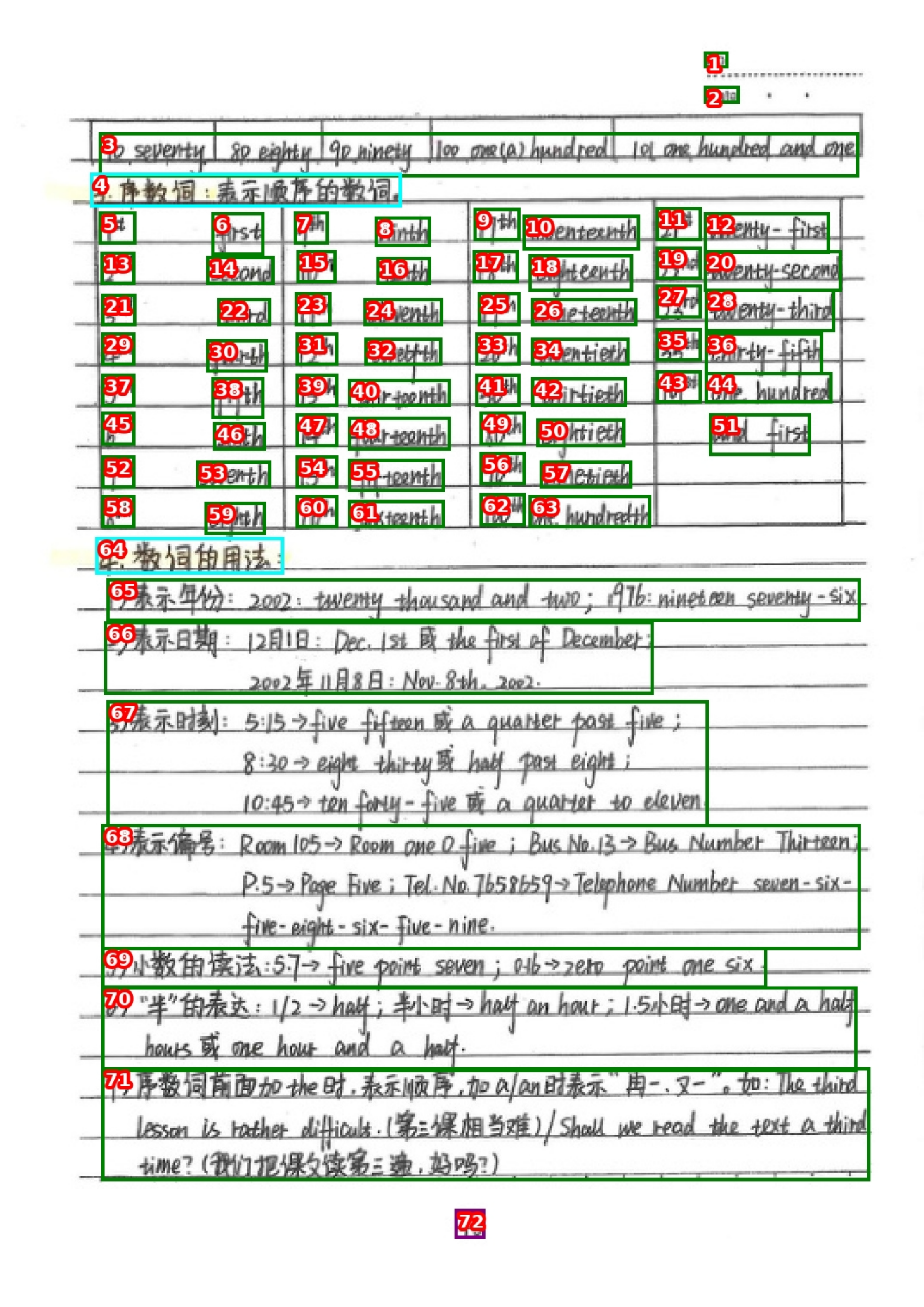}}
\end{minipage}

\caption{ \textbf{Effect of Removing the Recognition task (4).}
In the upper-left corner, the red numbers indicate the intended reading order of the paragraphs.
(b) Our unified model detects the entire word table as a coherent structure, preserving its layout and correctly following the reading sequence.  
(c) Without recognition supervision, the model breaks the table into individual word elements, drawing separate boxes for each token. Moreover, it outputs these elements in a horizontal row-by-row sequence rather than the intended vertical order.  
This highlights the critical role of recognition in capturing high-level structural semantics.
}

\label{fig:ablation_recognition4}
\end{figure}

As argued in the main text, while ablating text recognition $(M_\text{-Rec})$
may lead to a marginal improvement in the isolated detection metric, it comes at the cost of semantic understanding and end-to-end performance. Qualitative comparisons in Figures~\ref{fig:ablation_recognition1}--\ref{fig:ablation_recognition4} visually highlight this degradation. Specifically, the absence of recognition supervision results in detections misaligned with the logical structure of the text, causing fragmented bounding boxes and incorrect reading orders.

\begin{figure}[ht]
\centering

\begin{minipage}{0.32\linewidth}
\centering
\textbf{(a) Ground Truth} \\[4pt]
\fbox{\includegraphics[width=0.9\linewidth]{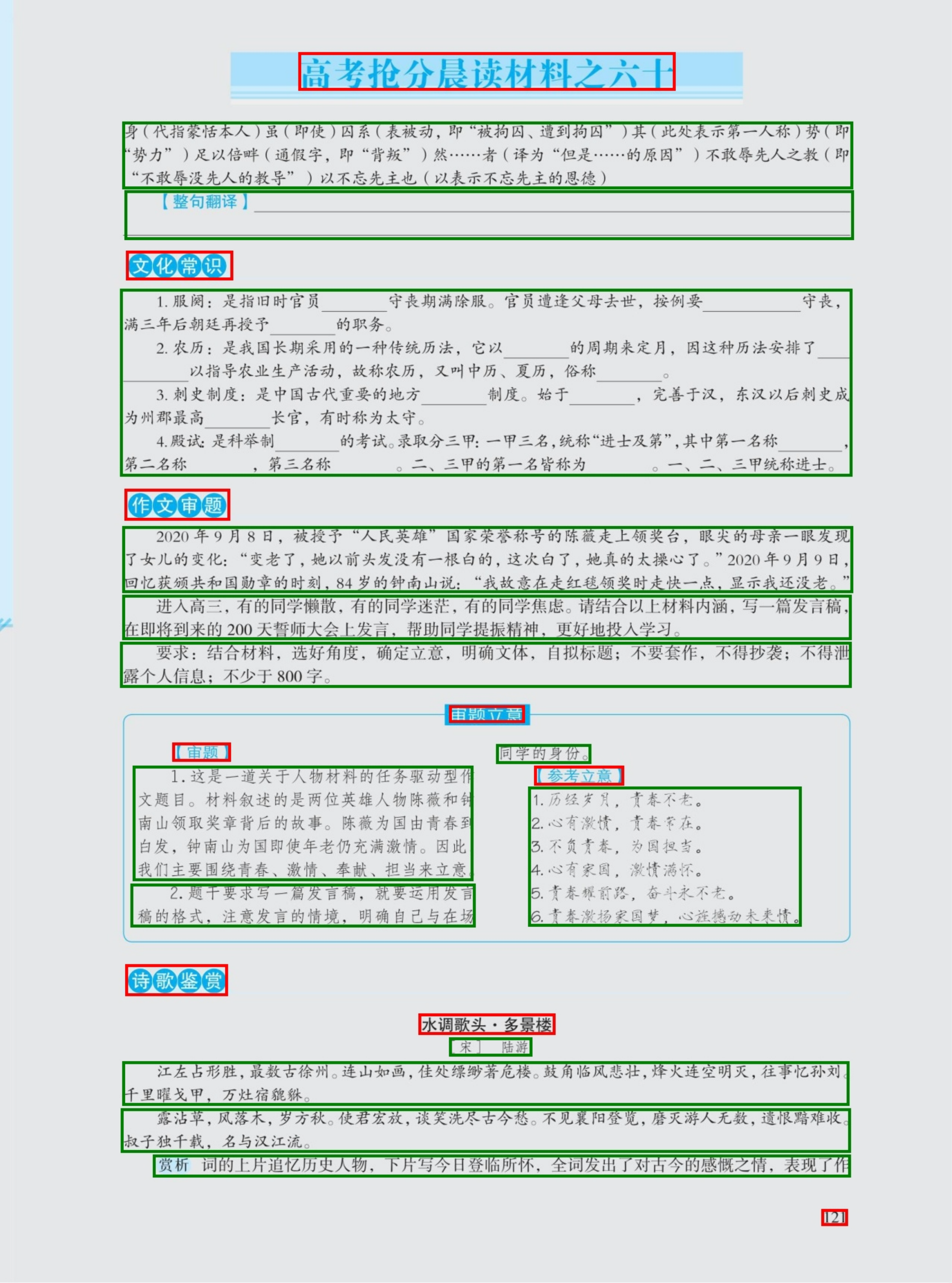}}
\end{minipage}
\hfill
\begin{minipage}{0.32\linewidth}
\centering
\textbf{(b) Our Unified Model} \\[4pt]
\fbox{\includegraphics[width=0.9\linewidth]{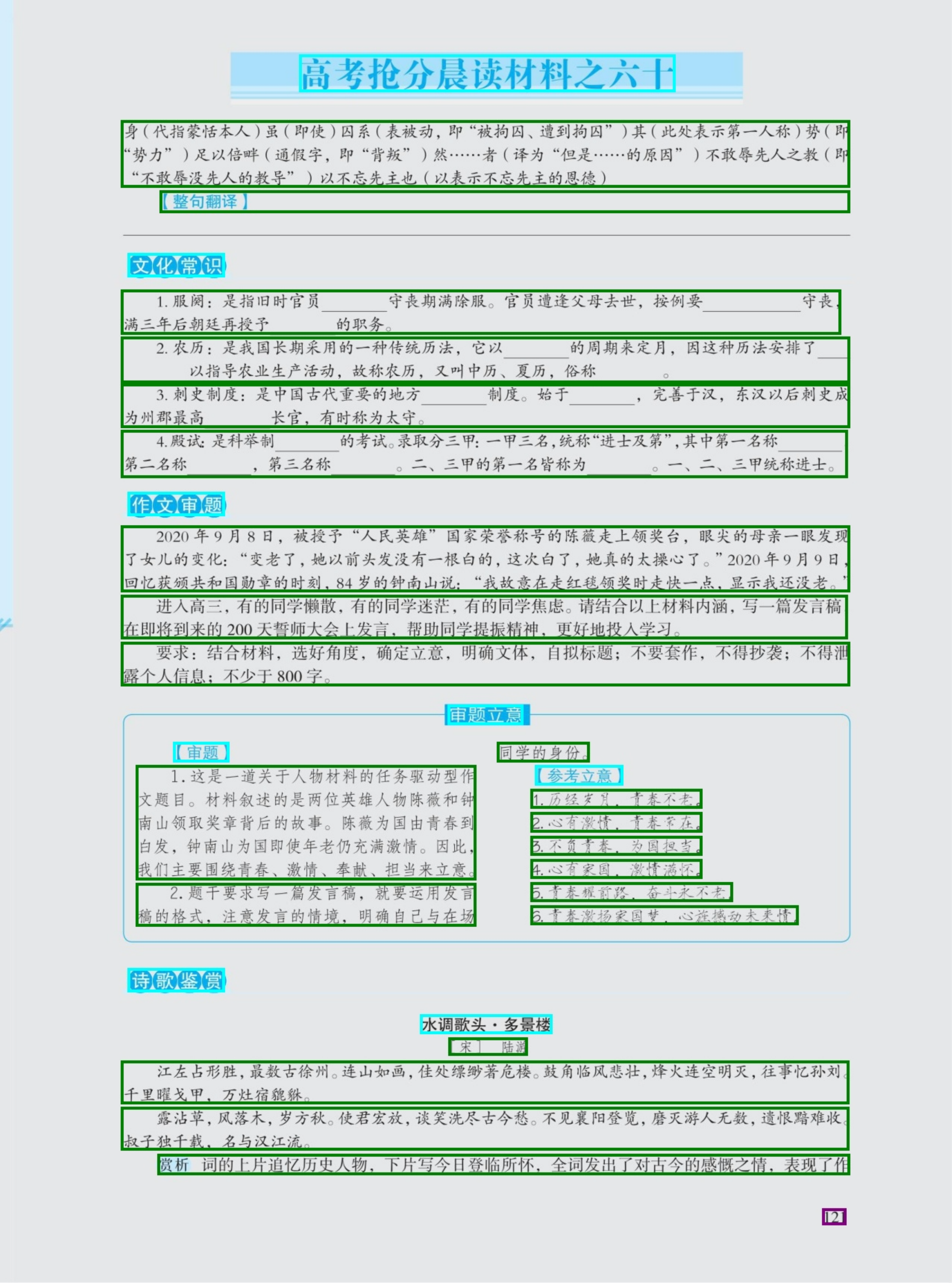}}
\end{minipage}
\hfill
\begin{minipage}{0.32\linewidth}
\centering
\textbf{(c) M$_{\text{-RO}}$ (without reading order)} \\[4pt]
\fbox{\includegraphics[width=0.9\linewidth]{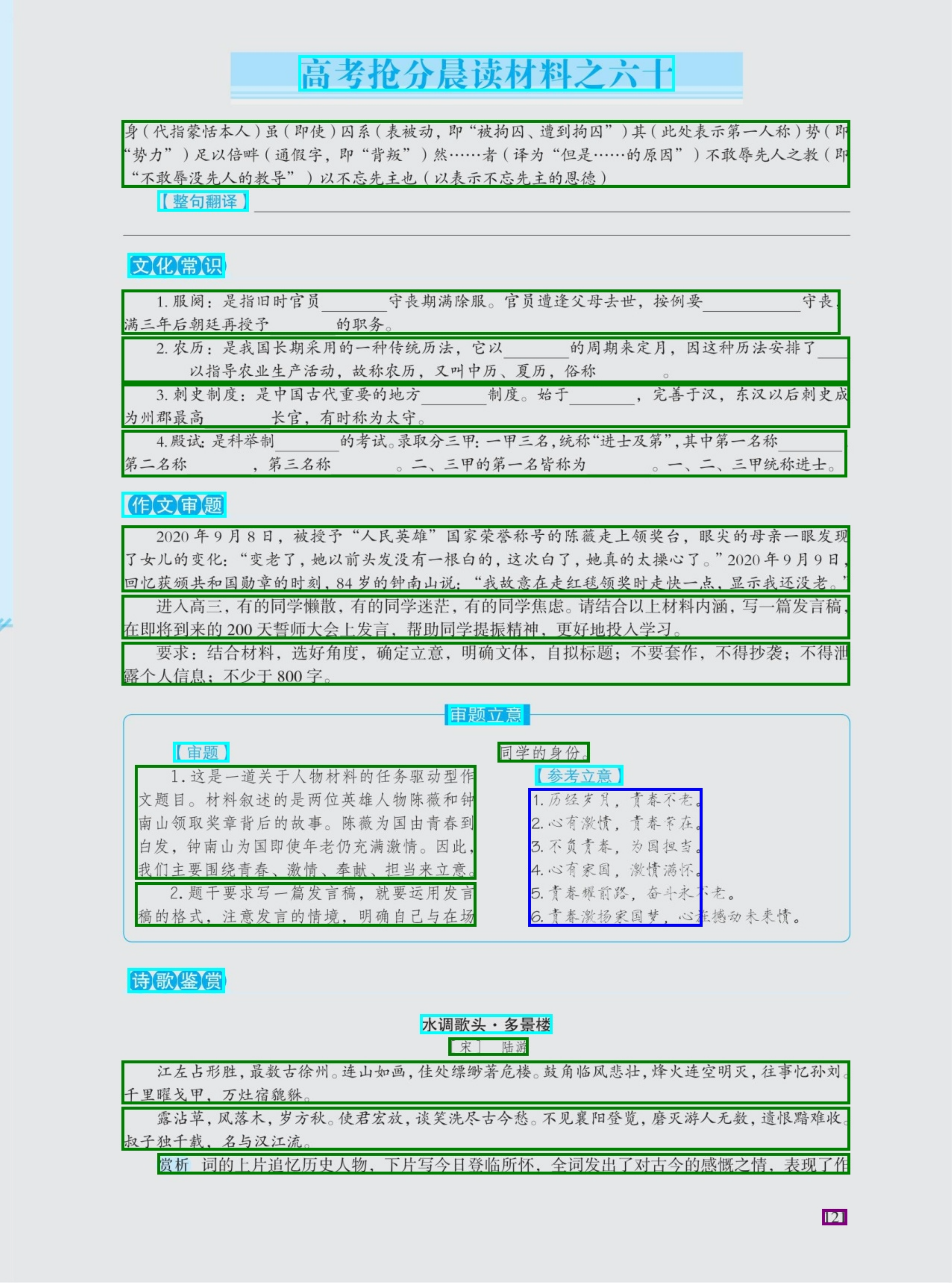}}
\end{minipage}

\caption{ \textbf{Effect of Removing reading order (1).}
(b) our unified model generates structurally coherent detection boxes that accurately encapsulate text groups.
(c) Upon removing reading-order supervision, the generated detection boxes become misaligned. This underscores that proper reading-order supervision is vital for achieving accurate detection within a unified framework.
}

\label{fig:readingorder_case1}
\end{figure}

\begin{figure}[ht]
\centering

\begin{minipage}{0.32\linewidth}
\centering
\textbf{(a) Ground Truth} \\[4pt]
\fbox{\includegraphics[width=0.9\linewidth]{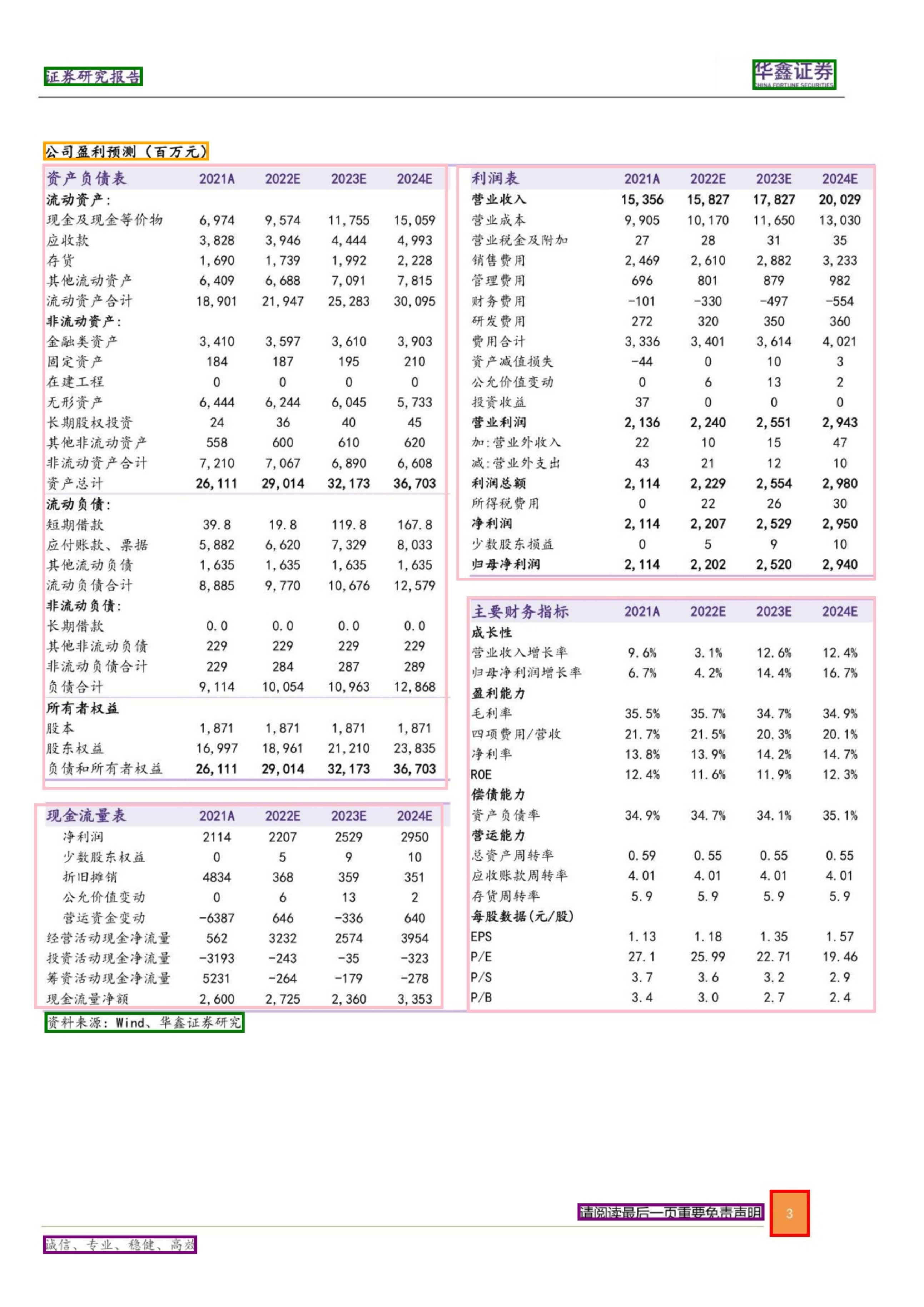}}
\end{minipage}
\hfill
\begin{minipage}{0.32\linewidth}
\centering
\textbf{(b) Our Unified Model} \\[4pt]
\fbox{\includegraphics[width=0.9\linewidth]{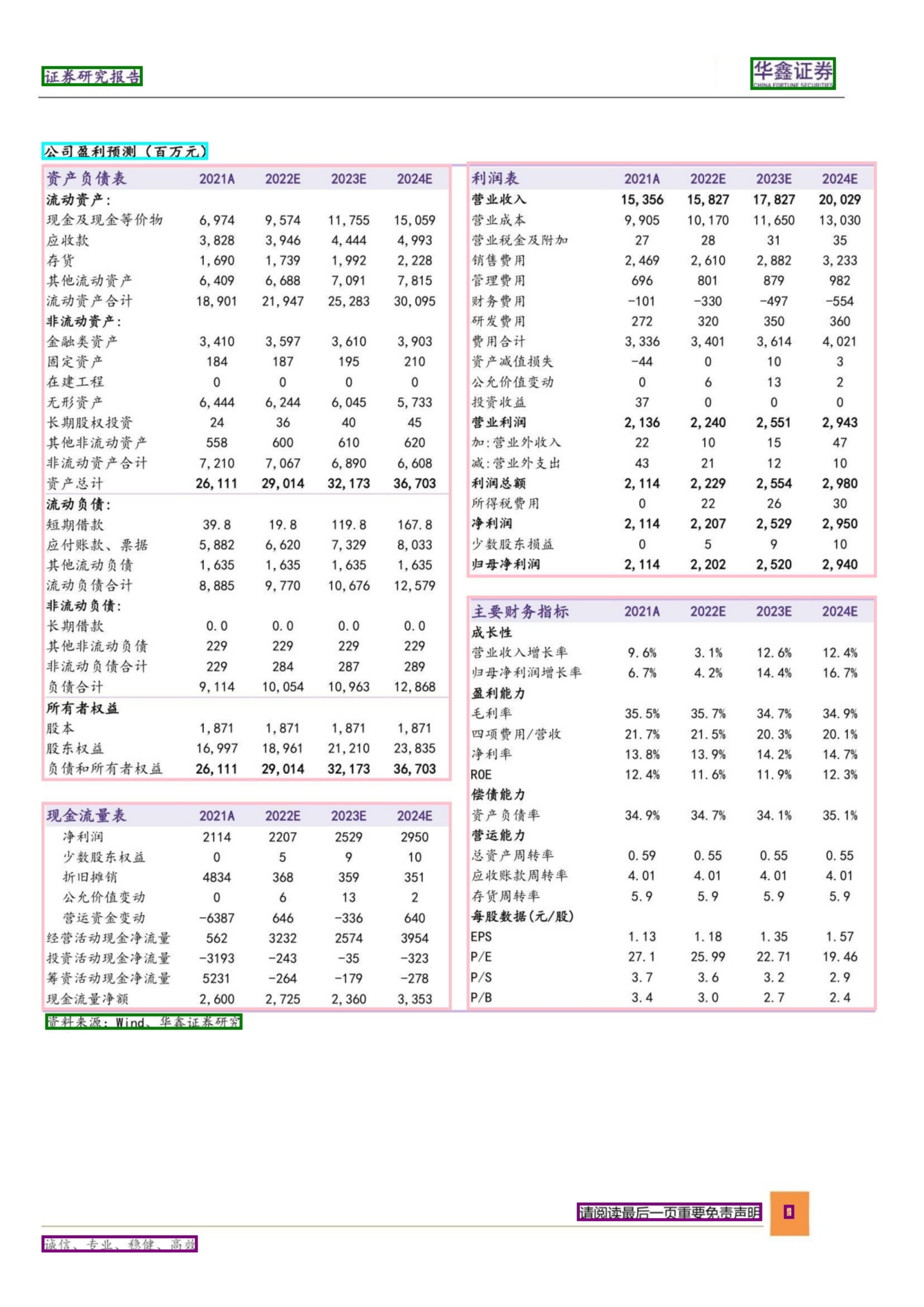}}
\end{minipage}
\hfill
\begin{minipage}{0.32\linewidth}
\centering
\textbf{(c) M$_{\text{-RO}}$ (without reading order)} \\[4pt]
\fbox{\includegraphics[width=0.9\linewidth]{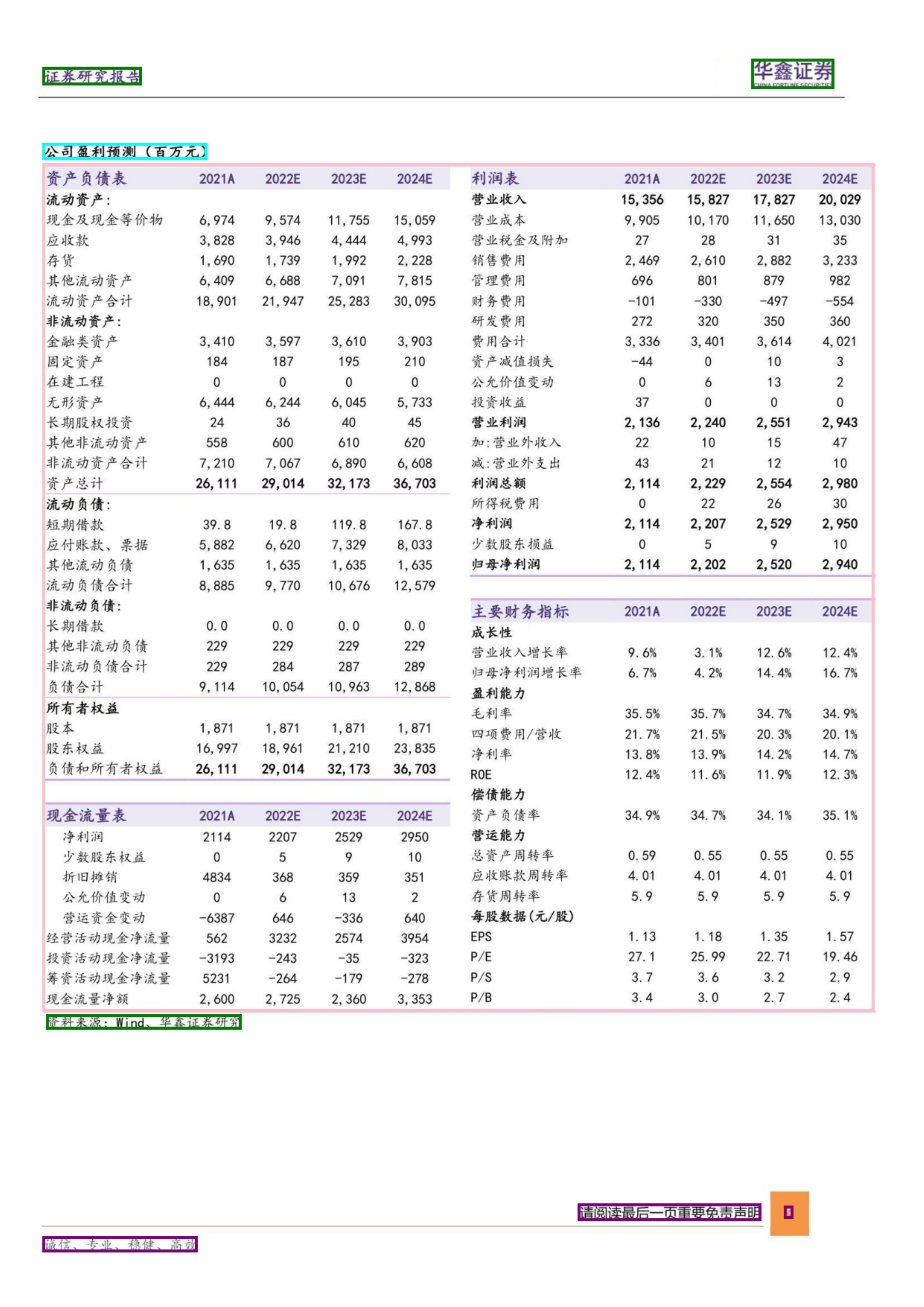}}
\end{minipage}

\caption{ \textbf{Effect of Removing reading order (2).}
The page contains four separate tables, each representing independent content blocks in the document layout.  
(b) Our unified model, trained with reading-order supervision, detects each table as an individual structured region, accurately preserving the separation between them.  
(c) Without correct reading-order supervision, the model fails to perceive these boundaries, incorrectly grouping all four tables into a single large bounding box.  
Clearly, lacking correct reading-order guidance significantly degrades the model’s structural understanding and its ability to distinguish separate entities.
}

\label{fig:readingorder_case2}
\end{figure}

We further present two qualitative examples obtained by ablating the reading order supervision $(M_\text{-RO})$. 
As shown in Figure~\ref{fig:readingorder_case1}, the absence of this guidance results in misaligned bounding regions. Furthermore, Figure~\ref{fig:readingorder_case2} reveals a critical failure where the model cannot discern four independent tables, erroneously aggregating them into a single, spanning box.

\section{dots.ocr as a Data Engine for VLM training}
\label{sec:data_generate}

As we posited in the main paper, the ultimate value of dots.ocr extends beyond its excellent performance on document analysis benchmarks. Its true significance lies in serving as a powerful data engine to fuel the next generation of VLMs. Here, we elaborate on this premise and substantiate its broader implications.

\begin{figure*}[ht]
  \centering
  \includegraphics[width=1.0\linewidth]{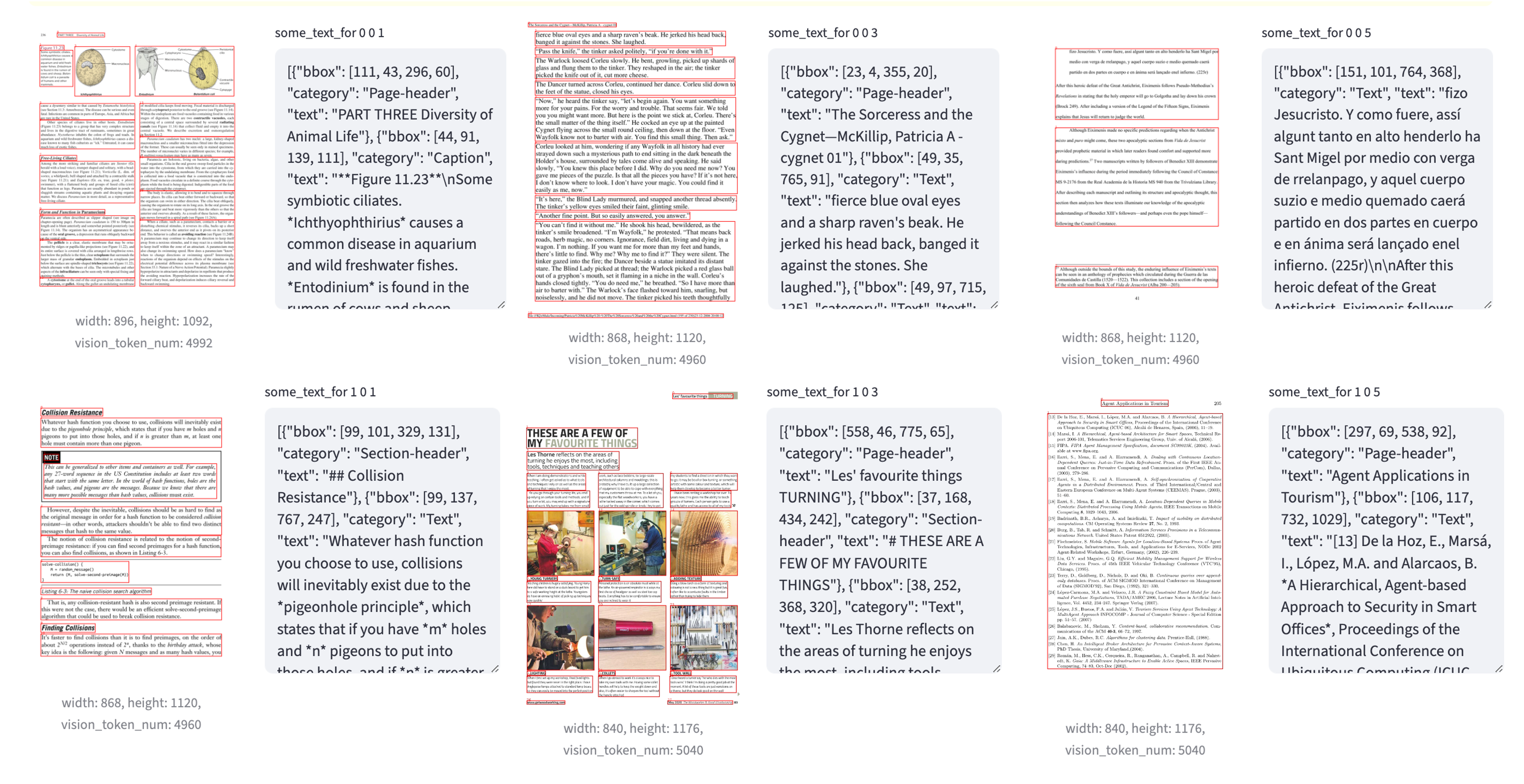}
    \caption{\textbf{Grounding-enhanced OCR enabled by dots.ocr.} The system accurately extracts text regions and their coordinates from the document, enabling the formulation of fine-grained vision-language grounding tasks.}
   \label{fig:grounding}
\end{figure*}

\begin{figure*}[ht]
  \centering
  \includegraphics[width=1.0\linewidth]{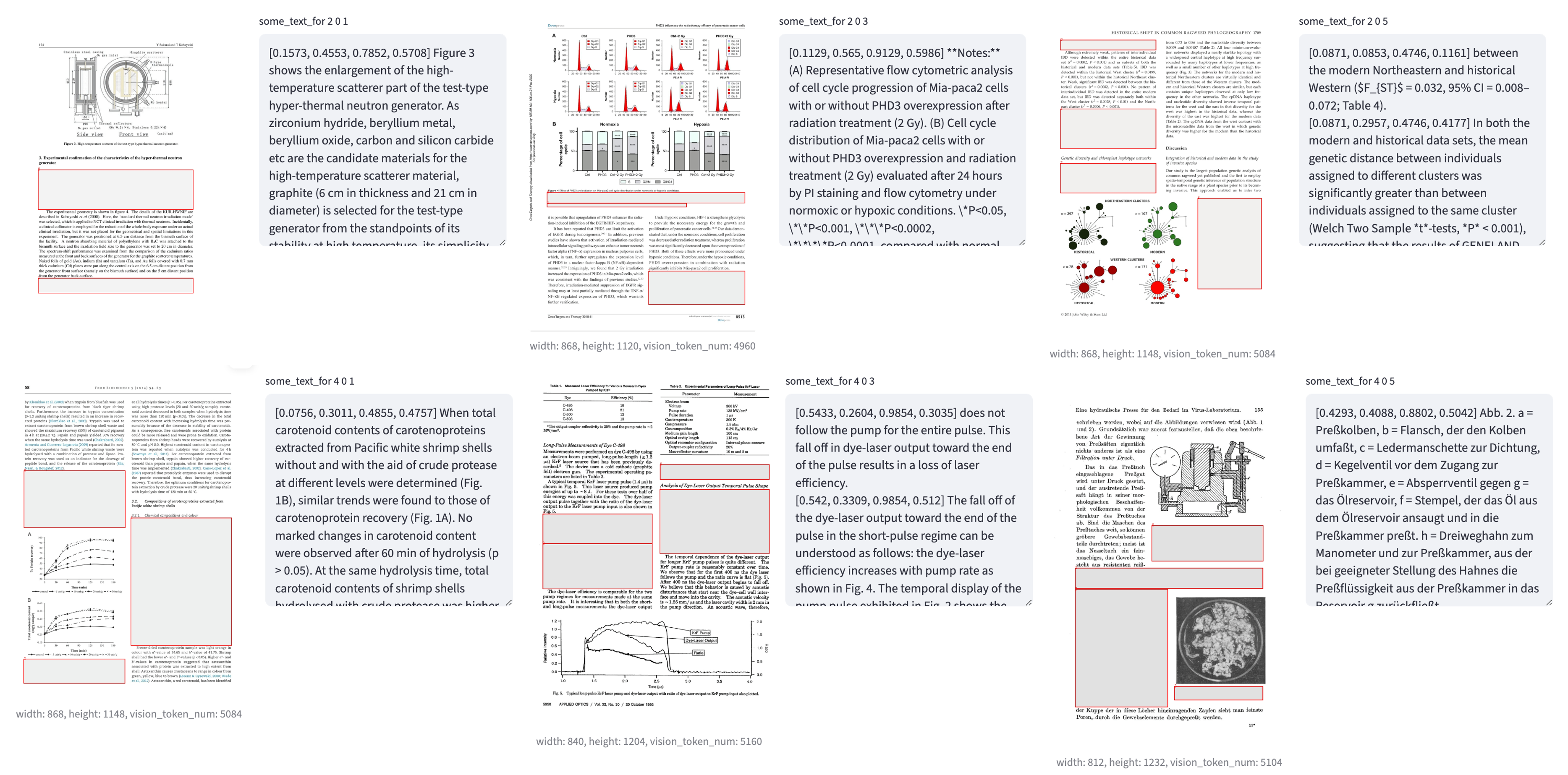}
    \caption{\textbf{Illustration of text-inpainting tasks generated with dots.ocr.} The system automatically masks text segments with precise bounding boxes, producing high-fidelity training data for models to predict missing content based on surrounding visual and textual context.}
   \label{fig:text_inpainting}
\end{figure*}

\begin{figure*}[ht]
  \centering
  \includegraphics[width=1.0\linewidth]{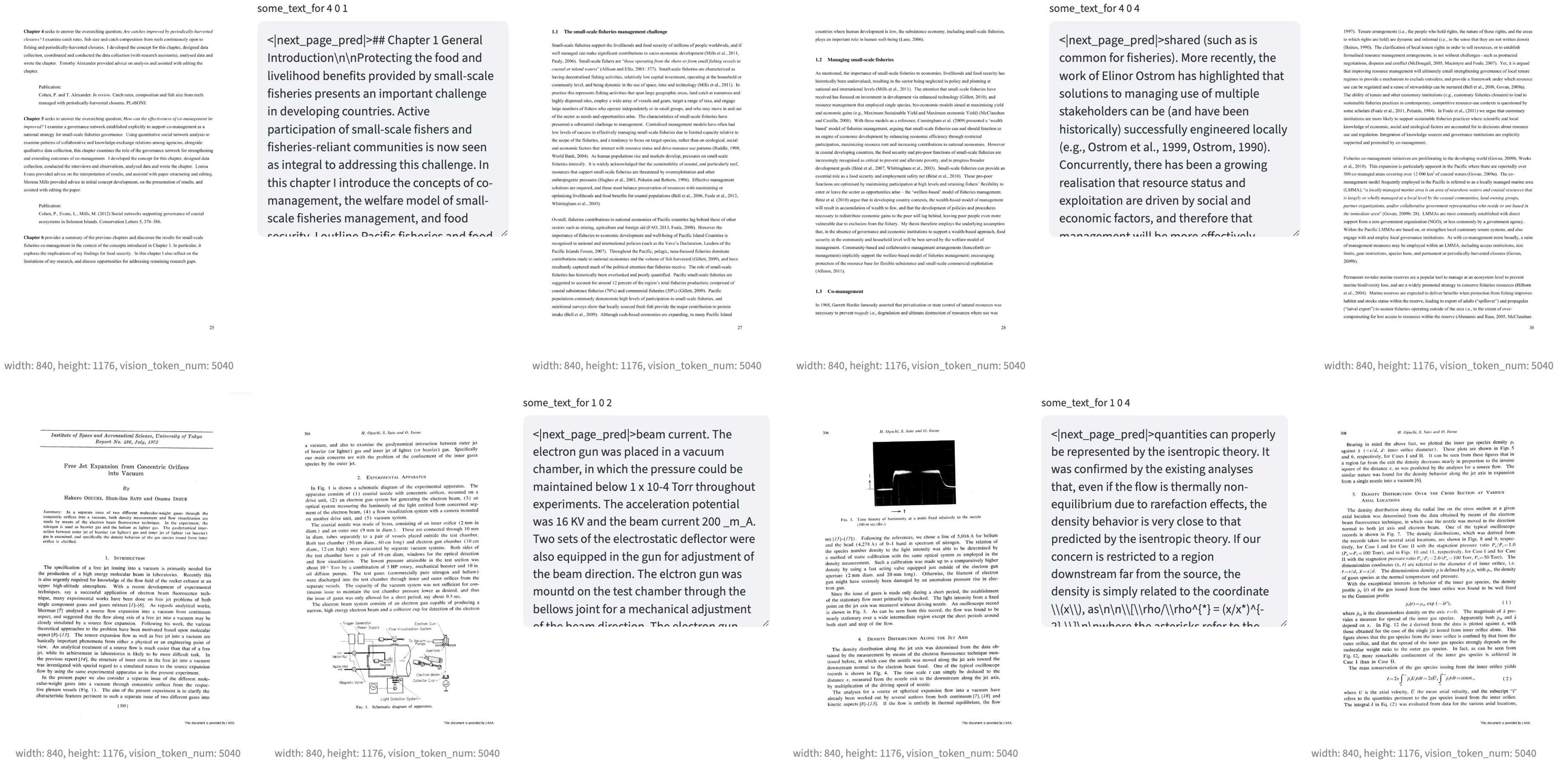}
    \caption{\textbf{Examples of next-page prediction pairs generated by dots.ocr.} For each current page, the corresponding next page is provided, allowing models to learn narrative flow and long-range document context.}
   \label{fig:next_page_predict}
\end{figure*}

We recognize that documents serve as a vast, untapped reservoir of human knowledge. dots.ocr acts as a pivotal tool to unlock this wealth, enabling the efficient extraction of high-quality multimodal information for VLM training. Specifically, our system facilitates a diverse array of downstream tasks: for lengthy volumes such as books with thousands of pages, it extracts coherent long-context textual narratives; it provides precise bounding box annotations to enhance the model's visual grounding capabilities; it mines rich image-caption pairs interleaved within the text; and it supports advanced generative objectives, including visual text inpainting and next-page text prediction. Figures~\ref{fig:grounding}--\ref{fig:next_page_predict} visually exemplify these capabilities, showcasing the quality and diversity of the training data generated by our model.

\begin{figure*}[ht]
  \centering
  \includegraphics[width=1.0\linewidth]{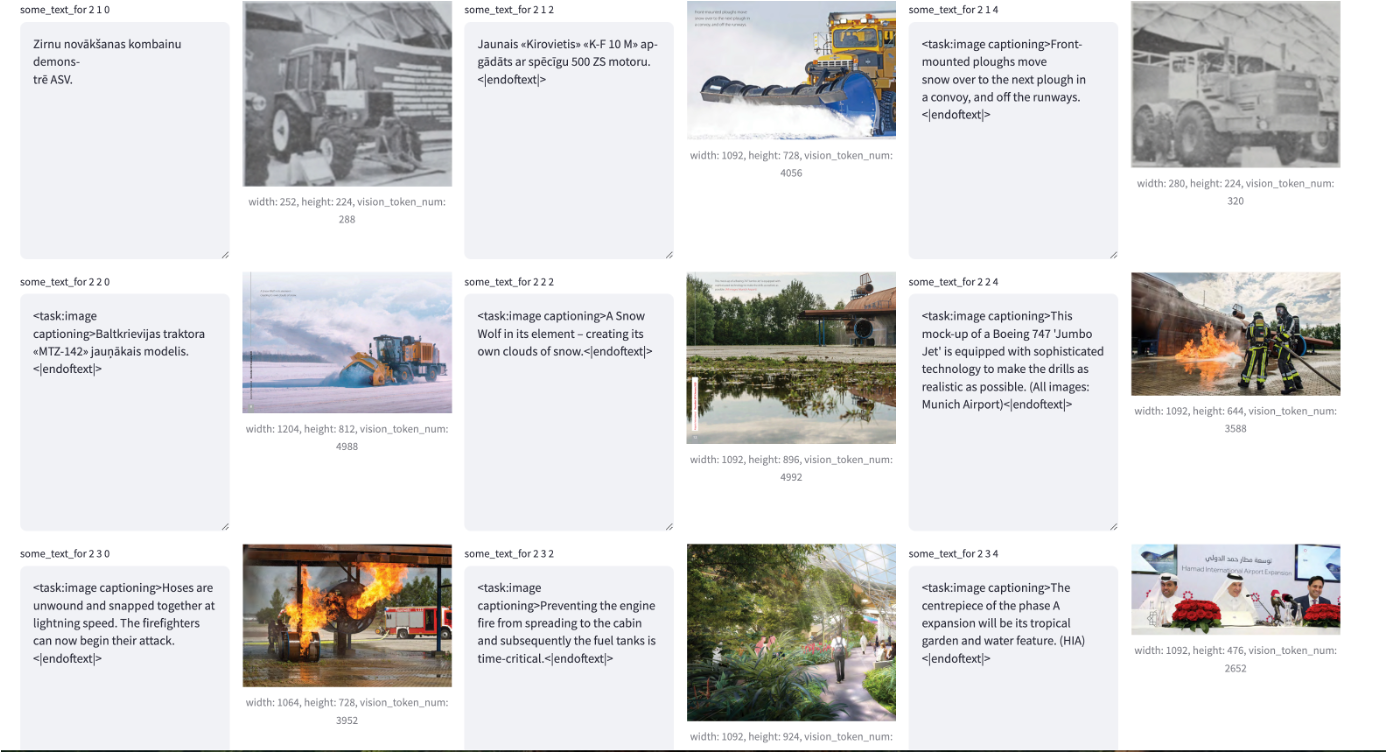}
    \caption{\textbf{Natural image-caption pairs extracted extracted by dots.ocr from diverse documents.} Our system effectively mines high-quality photographs and their corresponding descriptions from diverse document sources. These natural image-text pairs are crucial for enhancing the general visual understanding and cross-modal alignment capabilities of VLMs.}
   \label{fig:image_caption2}
\end{figure*}

\begin{figure*}[ht]
  \centering
  \includegraphics[width=1.0\linewidth]{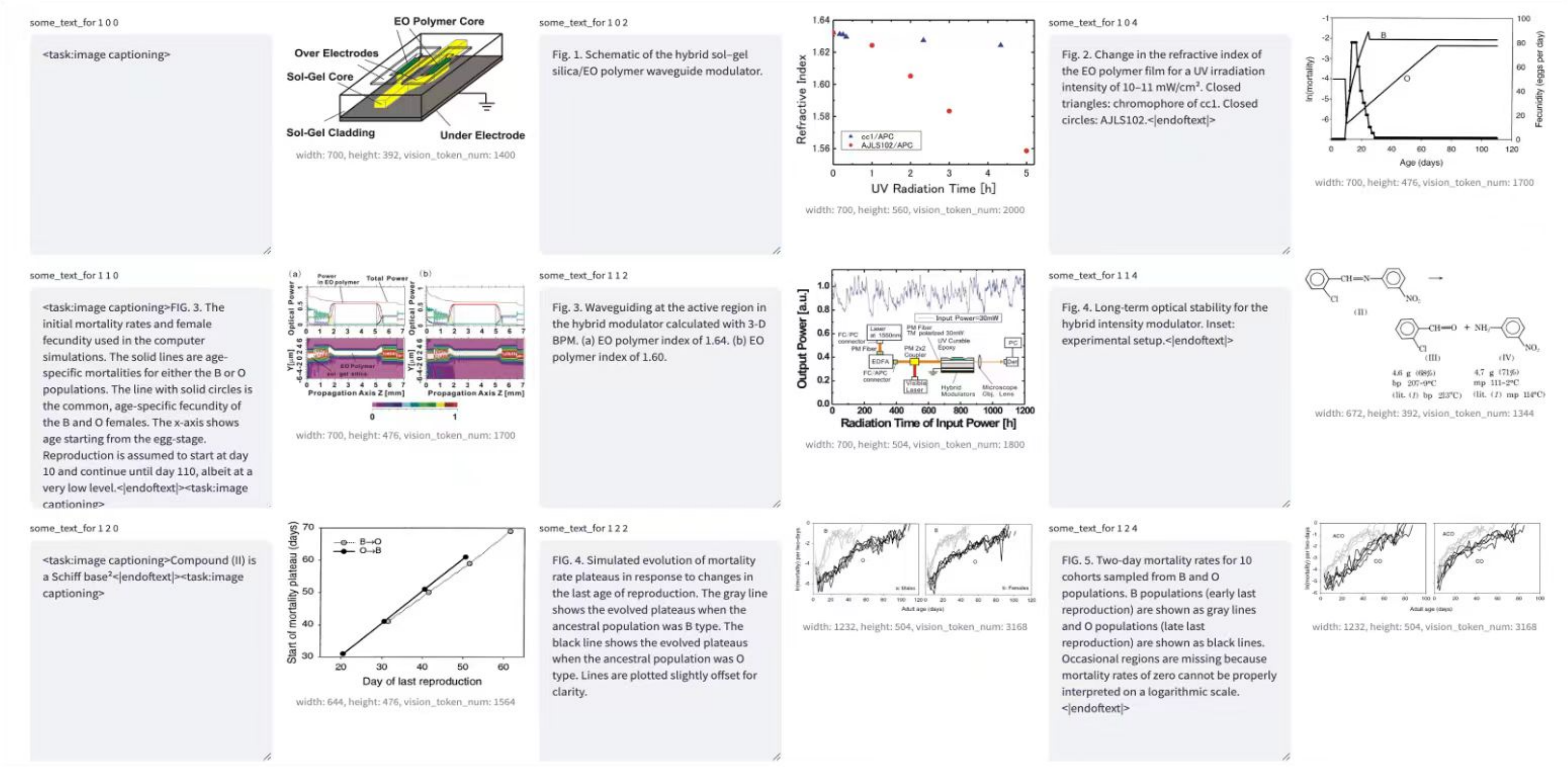}
    \caption{\textbf{Scientific figure-caption pairs extracted by dots.ocr.} Beyond natural images, our model accurately identifies and associates complex scientific charts, diagrams, and plots with their explanatory captions. This data is invaluable for training VLMs on domain-specific knowledge and fine-grained chart reasoning tasks.}
   \label{fig:image_caption2}
\end{figure*}

\end{document}